\documentclass{elsart}

\usepackage{times,mathptm}
\usepackage{times}
\usepackage{graphicx}
\usepackage{latexsym}
\usepackage{amssymb}

\newcommand{\shrink}[1]{}

\newcommand{\func}[1]{\textbf{function }\texttt{#1}}
\newcommand{\myendf}{\textbf{endfunction }}
\newcommand{\call}[1]{\texttt{#1}}

\newcommand{\mywhile}{\textbf{while}}
\newcommand{\myendw}{\textbf{endwhile}}
\newcommand{\mydo}{\textbf{do}}

\newcommand{\myif}{\textbf{if}}
\newcommand{\myapply}{\textbf{apply}}
\newcommand{\myuntil}{\textbf{until}}
\newcommand{\mythen}{\textbf{then}}

\newcommand{\myelseif}{\textbf{elseif}}

\newcommand{\mytrue}{\textbf{true}}
\newcommand{\myfalse}{\textbf{false}}

\newcommand{\myret}{\textbf{return}}

\begin{document}
\begin{frontmatter}

\title{A Logical Approach to Efficient Max-SAT solving\thanksref{label1}}
\thanks[label1]{This paper includes and extends preliminary work from \cite{Larrosa05a,Larrosa06}}

\author[label2]{Javier Larrosa}
\author[label3]{Federico Heras}
\author[label4]{Simon de Givry}
\address[label2]{\texttt{larrosa@lsi.upc.edu}}
\address[label3]{\texttt{fheras@lsi.upc.edu}\\
Universitat Politecnica de Catalunya\\
Barcelona, Spain}
\address[label4]{\texttt{degivry@toulouse.inra.fr}\\
INRA\\
Toulouse, France}

\begin{abstract}
Weighted Max-SAT is the optimization version of SAT and many important 
problems can be naturally encoded as such. Solving weighted Max-SAT
is an important problem from both a theoretical and a
practical point of view. In recent years, there has been considerable
interest in finding efficient solving techniques.
Most of this work focus on the
computation of good quality lower bounds to be used within a branch
and bound DPLL-like algorithm. Most often, these lower bounds are described in a
procedural way. Because of that, it is difficult to realize the {\em
logic} that is behind. 

In this paper we introduce an original framework for Max-SAT that stresses
the parallelism with classical SAT. 
Then, we extend the two basic SAT solving techniques: {\em search} and {\em inference}.
We show that 
many algorithmic {\em tricks} used in state-of-the-art Max-SAT solvers
are easily expressable in {\em logic} terms with our framework in a unified manner. 

Besides, we introduce an original search algorithm that performs a restricted
amount of {\em weighted resolution} at each visited node. 
We empirically compare our algorithm with a variety of solving alternatives on several benchmarks.
Our experiments, which constitute to the best of our knowledge the most comprehensive
Max-sat evaluation ever reported, show that our algorithm is generally
orders of magnitude faster than any competitor.
\end{abstract}

\end{frontmatter}

\section{Introduction}

Weighted Max-SAT is the optimization version of the SAT problem and many important problems can be naturally expressed as such. They include academic problems such as {\em max cut} or {\em max clique}, as well as real problems in domains like {\em routing} \cite{Xu02}, {\em bioinformatics} \cite{Strickland05}, {\em scheduling} \cite{vasquez01}, {\em probabilistic reasoning} \cite{Park02}, {\em electronic markets} \cite{sandholm99}. In recent years, there has been a considerable effort in finding efficient exact algorithms. These works can be divided into theoretical \cite{Bansal99,Niedermeier00,Chen2002} and empirical \cite{Givry03,Shen04,Zhang05,Alsinet05}. A common drawback of all these algorithms is that albeit the close relationship between SAT and Max-SAT, they cannot be easily described with {\em logic} terminology. For instance, the contributions of \cite{Givry03,Shen04,Zhang05,Alsinet05} are good quality lower bounds to be incorporated into a {\em depth-first branch and bound} procedure. These lower bounds are mostly defined in a procedural way and it is very difficult to see the {\em logic} that is behind the execution of the procedure. This is in contrast with SAT algorithms where the solving process can be easily decomposed into atomic logical steps.
%(e.g. search algorithms backtrack when they derive an empty clause $\Box$, 
%unit propagations is a sequence of resolution steps of a special kind, etc.)

\shrink{
, meaning that the current subproblem is examined in the search of disjunct local inconsistencies (i.e, a subset of conflicting clauses) and the lower bound is the sum of all the identified inconsistencies.
There are two drawbacks for such an approach.
The first one is conceptual, because it is hard to see the {\em logic} that is behind the execution of an algorithm. 
The second one is practical, because these algorithms usually require to compute the lower bound from scratch at 
each visited node. This is inefficient, because the subproblem associated to a search node may share many similarities
with its parent, so part of the work could have been re-used.
Observe that this is not the case in classical SAT, where search algorithms simplify the formula associated to each visited node, transforming it into a presumably simpler equivalent one. If the simplification raises an explicit contradiction, the algorithm backtracks. Else, the computational effort of the simplification is still beneficial, because it is used in the subsequent search.
}

In this paper we introduce an original framework for (weighted) Max-SAT in which the notions of {\em upper} and {\em lower bound} are incorporated into the problem definition. Under this framework classical SAT is just a particular case of Max-SAT, and the main SAT solving techniques can be naturally extended. In particular, we extend the basic simplification rules (for example, {\em idempotency}, {\em absorption}, {\em unit clause reduction}, {\em etc}) and introduce a new one, {\em hardening}, that does not make sense in the SAT context. We also extend the two fundamental SAT algorithms: DPLL (based on {\em search}) and DP (based on {\em inference}). We also show that the complexity of the extension of DP is exponential on the formula's {\em induced width} (which is hardly a surprise, since this is also the case of other inference algorithms for graphical models \cite{Bertele72,Dechter}). Interestingly, our resolution rule includes, as special cases, many techniques spread over the recent Max-SAT literature. One merit of our framework is that it allows to see all these techniques as inference rules that {\em transform} the problem into an equivalent simpler one, as it is customary in the SAT context. 

The second contribution of this paper is more practical. We introduce an original search algorithm that incorporates three different forms of resolution at each visited node: {\em neighborhood resolution}, {\em chain resolution} and {\em cycle resolution}. Our experimental results on a variety of domains indicate that our algorithm is orders of magnitude faster than its competitors. This is especially true as the ratio between the number of clauses and the number of variables increases. Note that these are typically the hardest instances for Max-SAT. Our experiments include random weighted and unweighted Max-SAT instances, random and structured Max-one problems, random Max-cut problems, random and structured Max-clique problems and combinatorial auctions.

Some of the ideas presented in this paper have strong connections to different techniques recently developed in the WCSP field \cite{Meseguer06}. Especially significant is the connection with {\em local consistency} \cite{Cooper03,Cooper04,Larrosa03,Larrosa04,deGivry05} and {\em variable elimination} \cite{Bertele72,Dechter99,larrosa03b}.

The structure of the paper is as follows: In Section 2 we review SAT terminology. In Section 3 we present Max-SAT and introduce our framework. In Section 4 we extend from SAT to Max-SAT the essential solving techniques. Section 5 summarizes in a unified way special forms of resolution that can be used to simplify Max-SAT formula. Section 6 describes our solver. Section 7 reports our experimental work, which corroborate the efficiency of our solver compared to other state-of-the-art solving alternatives. 
Finally, Section 8 concludes and points out directions of future work.

\section{Preliminaries on SAT}

In the sequel $X=\{x_1,x_2,\ldots,x_n\}$ is a set of boolean
variables. 
A {\em literal} is
either a variable $x_i$ or its negation $\bar{x_i}$. The variable to which literal $l$ refers is noted $var(l)$ (namely, $var(x_i)=var(\bar{x}_i)=x_i$).
If variable $x_i$ is assigned to {\em true}
literal $x_i$ is satisfied and literal $\bar{x}_i$ is falsified. Similarly,
if variable $x_i$ is instantiated to {\em false}, literal $\bar{x}_i$ is satisfied and literal $x_i$ is
falsified.  
An assignment is {\em complete} if it
gives values to all the variables in $X$ (otherwise it is partial). 
A {\em clause} $C=l_1\vee l_2 \vee \ldots \vee l_k$ is a disjunction of literals such that 
$\forall_{1\leq i,j\leq k,\ i\neq j}\ \ var(l_i)\neq var(l_j)$. It is customary to think of a clause
as a set of literals, which allows to use the usual set operations. If $x\in C$ (resp. $\bar{x}\in C$) we say that $x$ appears in the clause with positive (resp. negative) sign.
The size of a clause, noted $|C|$, 
is the number of literals that it has. $var(C)$ is the set of variables that appear in $C$ (namely,
$var(C)=\{var(l)|\ l\in C\}$).
An assignment satisfies a clause iff it satisfies one or more of its 
literals. Consequently, the empty clause, noted $\Box$, cannot be satisfied. 
Sometimes it is convenient to think of clause $C$ as its equivalent $C \vee \Box$.
A logical formula $\mathcal{F}$ in
{\em conjunctive normal form} (CNF) is a conjunction of different clauses, normally expressed as a set.
A satisfying complete assignment is called a {\em
model} of the formula. Given a CNF formula, the SAT problem
consists in determining whether there is any model for it or not.
The empty formula, noted $\emptyset$, is trivially satisfiable. 
A formula containing the empty clause is trivially unsatisfiable and we say that it contains an {\em explicit contradiction}.

\subsection{Graph concepts\cite{Rish00}}

\begin{figure}[t]
        \centering
	\includegraphics[width=8cm]{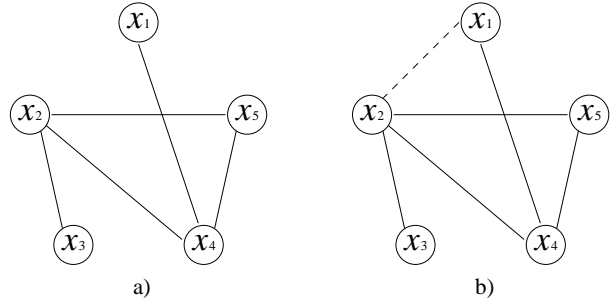}
	\caption{On the left, a graph $G$. On the right, the induced graph $G_d^*$ where $d$ is the lexicographic order.}
\label{p_graph1}
\end{figure}

The structure of a CNF formula $\mathcal{F}$ can be described by its {\em interaction graph} $G(\mathcal{F})$
containing one vertex associated to each boolean variable. There is an edge for each pair of vertices that
correspond to variables appearing in the same clause.
Given a graph $G$ and an ordering of its vertices $d$, the {\em parents} of a node $x_i$ is the set of vertices connected to $x_i$ that precede $x_i$ in the ordering. The {\em width} of $x_i$ along $d$ is the number of parents that it has. The {\em width of the graph} along $d$, denoted $w_d$, is the maximum width among the vertices.

The {\em induced graph} of $G(\mathcal{F})$ along $d$, denoted $G^*_d(\mathcal{F})$, is obtained as follows: The vertices of $G$ are processed from last to first along $d$. When processing vertex $x_i$, we connect every pair of unconnected parents. The {\em induced width} of $G$ along $d$, denoted $w^*_d$, is the width of the induced graph. 
The induced width (also known as {\em tree-width}, {\em k-tree number} or the {\em dimension of the graph}) is a measure of how far a graph is from acyclicity and it is a fundamental structural parameter in the characterization of many combinatorial algorithms. Computing the ordering $d$ that provides the minimum induced width is an NP-hard problem \cite{Arnborg85}.

\begin{exmp}
Consider the formula $\mathcal{F}=\{\bar{x}_1\vee x_4, x_1\vee x_4,x_2\vee x_3, x_2\vee x_4, x_2\vee \bar{x_5}, x_4\vee x_5\}$. Its interaction graph $G(\mathcal{F})$ is depicted in Figure \ref{p_graph1} ($a$). The induced graph $G^*_d$ along the lexicographical order is depicted in Figure \ref{p_graph1} ($b$). Dotted edge is the only new edge with respect the original graph. When processing node $x_5$, no new edges are added, because the parents of $x_5$ are already connected. When processing node $x_4$, the edge connecting $x_2$ and $x_1$ is added because both variables are parents of $x_4$ and they were not connected. When processing $x_3$, $x_2$ and $x_1$, no new edges are added. The induced width $w^*_d$ is $2$ because nodes $x_5$ and $x_4$ have width $2$ (namely, they have two parents) in the induced graph.
\end{exmp}

\subsection{SAT algorithms}

CNF formulas can be simplified using equivalences or reductions. Well known equivalences are
{\em idempotency} $C\wedge C \equiv C$,
{\em absorption} $C\wedge (C\vee B) \equiv C$ or
{\em unit clause reduction} $l \wedge (\bar{l}\vee C) \equiv l \wedge C$. A well known reduction is the 
{\em pure literal rule} which says that if there is a variable such that it only occurs in either positive or negative form, all clauses mentioning it can be discarded from the formula. 
Simplifications can be applied until quiescence. The assignment of {\em true} (resp. {\em false}) to variable $x$ in $\mathcal{F}$ is noted $\mathcal{F}[x]$ (resp. $\mathcal{F}[\bar{x}]$) and produces a new formula in which all clauses containing $x$ (resp. $\bar{x}$) are eliminated from the formula, and $\bar{x}$ (resp. $x$) is removed from all clauses where it appears. Note that $\mathcal{F}[l]$ can be seen as the addition of $l$ to the formula and the repeated application of unit clause reduction followed by the pure literal rule.

\begin{figure}

\begin{tabbing}
mm\=mm\=mm\=mm\=mm\=mm\=mm\=mm\=\kill
\func{DPLL}($\mathcal{F}$) \myret\ boolean \\
1.\> $\mathcal{F} :=$ \call{Simplify}($\mathcal{F}$)\\
2.\> \myif\ $\mathcal{F}=\emptyset$ \ \mythen\ \myret\ $\mytrue$\\
3.\> \myif\ $\mathcal {F}=\{\Box\}$\ \mythen\ \myret\ $\myfalse$\\
4.\> $l:=$\call{SelectLiteral}($\mathcal{F}$)\\
5.\> \myret\ \call{DPLL}($\mathcal{F}[l]$) $\vee$ 
          \call{DPLL}($\mathcal{F}[\bar{l}]$)\\
\myendf\\
\end{tabbing}

\caption{DPLL is a search algorithm. It returns $true$ iff $\mathcal{F}$ is satisfiable. \label{DPLL}}
\end{figure}

Algorithms for SAT can be roughly divided into {\em search} and {\em inference}.
The most popular search algorithm and the starting point of most state-of-the-art SAT solvers
was proposed in \cite{Davis62} and is usually called {\em Davis Putnam Logemann Loveland} (DPLL).
Figure \ref{DPLL} provides a recursive description. 
First, DPLL simplifies its input (line 1).
If the resulting formula is empty, it reports success (line 2). 
If the resulting formula is a contradiction, it reports failure (line 3). 
Else it selects a literal $l$ (line 4) and sequentially assigns the formula with $l$ and $\bar{l}$ (line 5).

We say that two clauses $x\vee A, \bar{x}\vee B \in \mathcal{F}$ {\em clash} iff $A\vee B$ is not a tautology (namely, $\forall_{l\in A}\ \bar{l}\notin B$) and, is not absorbed (namely, $\forall_{C\in \mathcal{F}}\ C\not\subseteq A\vee B$).
The {\em resolution} rule,
$\{x\vee A, \bar{x}\vee B\} \equiv \{x\vee A, \bar{x}\vee B, A\vee B\}$, is applied to clashing clauses and
is central to inference algorithms. 
Variable $x$ is called the {\em clashing variable} and $A\vee B$ is called the {\em resolvent}. Resolution, which is sound and complete, adds to the formula (i.e, makes explicit) an implicit relation between $A$ and $B$.  
Note that unit clause reduction is just a particular case of resolution.

\begin{figure}

\begin{tabbing}
mm\=mm\=mm\=mm\=mm\=mm\=mm\=mm\=\kill

\func{VarElim}($\mathcal{F},x_i$) \myret\ CNF formula \\
1.\> $\mathcal{B}:=\{C\in \mathcal{F}|\ x_i\in var(C)\}$\\
2.\> $\mathcal{F}:= \mathcal{F}-\mathcal{B}$\\
3.\> \mywhile\ $\exists\ x_i\vee A \in \mathcal{B}$\ \mydo\\
4\> \> $x_i\vee A:=$\call{PopClause}($\mathcal{B}$)\\
5.\> \> \mywhile\ $\exists_{\bar{x}_i\vee B\in \mathcal{B}}$ s.t. \call{Clash}($x_i\vee A,\bar{x}_i\vee B$)\ \mydo\\
6.\>  \> \>$\mathcal{F}:=\mathcal{F} \cup \{A\vee B\}$\\
7.\> \>\myendw\\
8.\> \myendw\\
%2.\> \mywhile\ $\exists \{x\vee A,\bar{x}\vee C\} \subseteq \mathcal{B}$ s.t. clash \mydo\\
%3\> \> $(x\vee A, \bar{x}\vee C):=$\call{SelectClashingClauses}($\mathcal{B}$)\\
%4.\>  \>$\mathcal{F}:=\mathcal{F} \cup \{A\vee C\}$\\
%5.\> \myendw\\
%5.\> \myuntil\ $quiescence$\\
9. \> \myret\ ($\mathcal{F}$)\\
\myendf\\

\func{DP}($\mathcal{F}$) \myret\ boolean \\
10.\> $\mathcal{F} :=$ \call{Simplify}($\mathcal{F}$)\\
11.\> \myif\ $\mathcal{F}=\emptyset$ \ \mythen\ \myret\ $\mytrue$\\
12.\> \myif\ $\mathcal {F}=\{\Box\}$\ \mythen\ \myret\ $\myfalse$\\
13.\> $x_i:=$\call{SelectVar}($\mathcal{F}$)\\
14.\> \myret\ \call{DP}(\call{VarElim}($\mathcal{F},x_i$))\\
\myendf\\
\end{tabbing}

\caption{DP is a pure inference algorithm. It returns $true$ iff $\mathcal{F}$ is satisfiable. \label{DP}}
\end{figure}

Two years before DPLL, Davis and Putnam proved that a restricted amount of resolution performed along some ordering of the variables is sufficient for deciding satisfiability.  The corresponding algorithm is noted DP \cite{Davis60,Rish00}. Figure \ref{DP} provides a recursive description. It eliminates variables one-by-one until it obtains the empty formula or achieves a contradiction. 
The heart of DP is Function \texttt{VarElim}. It eliminates variable $x_i$ from formula $\mathcal{F}$ while preserving its solvability. First, it computes the so-called {\em bucket} of $x_i$, noted $\mathcal{B}$, which contains the set of clauses mentioning the variable (line 1). All the clauses in the bucket are removed from the formula (line 2).
Next, it applies resolution restricted to the clauses in the bucket while pairs of clashing clauses exist. Resolvents are added to the formula (line 6). The correctness of DP is based on the fact that clauses added in line 6 keep the essential information contained in clauses removed in line 2.
Observe that the pure literal rule is just a special case of variable elimination in which no pair of clashing clauses exist, so the inner loop never iterates.

%\begin{thm}\cite{Davis60}
%Algorithm DP is correct
%\end{thm}

The following lemma shows how the complexity of eliminating a variable depends on the number of other variables that it interacts with,
\begin{lem}\cite{Rish00}
Let $\mathcal{F}$ be a CNF formula and $x_i$ one of its variables.
Let $n_i$ be the number of variables sharing some clause with $x_i$ in $\mathcal{F}$.
The space and time complexity of \texttt{VarElim}($\mathcal{F},x_i$) is $O(3^{n_i})$ and $O(9^{n_i})$, respectively.
\end{lem}

The following lemma shows how the induced graph $G^*_d(\mathcal{F})$ captures the evolution of the interaction graph $G(\mathcal{F})$ as variables are eliminated.
\begin{lem}\cite{Rish00}
Let $d$ denote the reverse order in which DP($\mathcal{F}$) eliminates variables.
The width of $x_i$ along $d$ in the induced graph $G(\mathcal{F})^*_d$ bounds above the number of variables sharing some clause with $x_i$ at the time of its elimination.
\end{lem}

Thus, the induced width captures the most expensive variable elimination.
The following theorem, which  follows from the two previous lemmas, characterizes the complexity of DP in terms of the induced width.
\begin{thm}\cite{Rish00}
Let $d$ denote the reverse order in which DP($\mathcal{F}$) eliminates variables. Let $w^*_d$ denote the induced width of $G(\mathcal{F})$ along $d$. The space and time complexity of DP($\mathcal{F}$) is $O(n\times 3^{w^*_d})$ and $O(n\times 9^{w^*_d})$, respectively.
\end{thm}

A consequence of the previous theorem is that the order in which DP eliminates variables may be crucial for the algorithm's complexity. As an example, consider a formula, whose interaction graph is a tree of depth $1$. If variables are eliminated in a top-down order, the cost may be exponential on $n$. If variables are eliminated in a bottom-up order, the cost is linear. In general, finding optimal elimination orderings is an NP-hard problem and approximate algorithms must be used. In practical applications,
DP is generally too space consuming and cannot be used \cite{Rish00}.
Nevertheless, resolution still plays an important practical role in combination with search: the
addition of restricted forms of resolution at each search node 
anticipates the detection of dead-ends and improves its performance
\cite{Gelder95,Rish00,Bacchus02,Drake02}. As we will show, the use of resolution is even more relevant in the Max-SAT context.

\section{(Weighted) Max-SAT}

When a boolean formula does not have any model, one may be
interested in finding a complete assignment with minimum number of
violated clauses. This problem is known as {\em (unweighted) Max-SAT}. Note that no repetition of
clauses is allowed and all clauses are equally important. The complexity of Max-SAT is $P^{NP[\log n]}$,
meaning that it can be solved with a logarithmic number calls to a NP oracle \cite{Papadimitriou}.

{\em Weighted Max-SAT} is an extension of Max-SAT. A {\em weighted clause} is a pair $(C,w)$ such that 
$C$ is a classical clause and $w$ is a natural number indicating the  
cost of its falsification.
A weighted formula in conjunctive normal form is a  
{\em set} of weighted clauses.
The {\em cost} of
an assignment is the sum of weights of all the clauses that it
falsifies. Given a weighted formula, {\em weighted} Max-SAT is the problem
of finding a complete assignment with minimal cost. We can assume all clauses in the formula
being different, since $(C,u), (C,w)$ can be replaced by $(C,u+w)$. Note that clauses with cost $0$ do not have any effect and can be discarded.
Weighted Max-SAT is more expressive than unweighted Max-SAT and its complexity, $P^{NP}$, is higher \cite{Papadimitriou} (it may require a linear number of calls to a SAT oracle). Since most Max-SAT applications require the expressiveness of weights, in this paper we will focus on weighted Max-SAT. In the following, when we say Max-SAT we will be referring to {\em weighted} Max-SAT.

\begin{exmp}
\shrink{Given a graph $G=(V,E)$, a {\em cut} is defined by a subset of vertices $U\subseteq V$. The size of a cut is the number of edges $(v_i,v_j)$ such that $v_i\in U$ and $v_j\in V-U$. {\em Max-CUT} is a NP-Hard problem. It consists on finding a cut of maximum size. It can be easily modeled with unweighted Max-SAT. We associate one variable $x_i$ to each graph vertex. Value {\em true} (respectively, {\em false}) indicates that vertex $v_i$ belongs to $U$ (respectively, to $V-U$). For each edge $(v_i,v_j)$, there are two clauses $x_i \vee x_j, \bar{x}_i \vee \bar{x}_j$. Given a complete assignment, the number of violated clauses is $|E|-S$ where $S$ is the size of the cut associated to the assignment. As a consequence, the optimal Max-SAT assignment represents the optimal Max-CUT. For instance, consider the graph in Figure \ref{p_graph1}($a$). The corresponding Max-SAT formula is $\mathcal{F}=\{(x_1 \vee x_2),(\bar{x}_1 \vee \bar{x}_2),(x_1 \vee x_3),(\bar{x}_1 \vee \bar{x}_3),(x_1 \vee x_4),(\bar{x}_1 \vee \bar{x}_4),(x_1 \vee x_5),(\bar{x}_1 \vee \bar{x}_5),(x_2 \vee x_3),(\bar{x}_2 \vee \bar{x}_3),(x_2 \vee x_4),(\bar{x}_2 \vee \bar{x}_4),(x_3 \vee x_4),(\bar{x}_3 \vee \bar{x}_4),(x_4 \vee x_5),(\bar{x}_4 \vee \bar{x}_5)\}$. The optimal assignment is $\{x_1=\mathbf{t},x_2=\mathbf{t},x_3=\mathbf{f},x_4=\mathbf{f},x_5=\mathbf{t}\}$ which violates 2 clauses. It indicates that the size of the Max-CUT is $S=8-2=6$.}

Given a graph $G=(V,E)$, a {\em vertex covering} is a set $U\subseteq V$ such that for every edge $(v_i,v_j)$ either $v_i\in U$ or $v_j\in U$. The size of a vertex covering is $|U|$. The {\em minimum vertex covering} problem is a well-known NP-Hard problem. It consists in finding a covering of minimal size. It can be naturally formulated as (weighted) Max-SAT. We associate one variable $x_i$ to each graph vertex. Value {\em true} (respectively, {\em false}) indicates that vertex $x_i$ belongs to $U$ (respectively, to $V-U$). There is a binary weighted clause $(x_i \vee x_j,u)$ for each edge $(v_i,v_j)\in E$, where $u$ is a number larger than $|V|$. It specifies that at least one of these vertices must be in the covering because there is an edge connecting them.  There is a unary clause $(\bar{x}_i,1)$ for each variable $x_i$, in order to specify that it is preferred not to add vertices to $U$. Note that different weights in unary and binary clauses are required to express the relative importance of each type of clauses.

Consider the minimum vertex covering of the graph in Figure \ref{p_graph1} ($a$). The Max-SAT encoding is $\mathcal{F}=\{(\bar{x}_1,1),(\bar{x}_2,1),(\bar{x}_3,1),(\bar{x}_4,1),(\bar{x}_5,1),(x_1 \vee x_4,5),(x_2 \vee x_3,5),(x_2 \vee x_4,5),(x_2 \vee x_5,5),(x_4 \vee x_5,5)\}$. The optimal assignment is $\{x_2=x_4=\mathit{true},x_1=x_3=x_5=\mathit{false}\}$ with cost $2$  that is equal to the size of the minimum vertex covering.

\end{exmp}

Next, we propose an alternative, although equivalent, definition for weighted Max-SAT that will be more convenient for our purposes. Given a weighted CNF formula, 
we assume the existence of a known upper bound $\top$ on the cost of an optimal
solution ($\top$ is a strictly positive natural number). This is done without 
loss of generality because, if a tight upper bound
is not known, $\top$ can be set to any number higher than the sum of weights of all the clauses. A  {\em
model} for the formula is a complete 
assignment with cost less than $\top$. An {\em optimal model} is a model of minimal cost.
Then, Max-SAT can be reformulated as 
the problem of finding an optimal model, if there is any. 
Observe that any weight $w\geq \top$ indicates that
the associated clause {\em must be necessarily satisfied}. Thus, we
can replace $w$ by $\top$ without changing the
problem. Thus, without loss of generality we 
assume all costs in the interval $[0..\top]$ and, following \cite{Larrosa02}, redefine
the {\em sum of costs} as,
$$a\oplus b=min \{a+b,\top\}$$ 
in order to keep the
result within the interval $[0..\top]$. A clause with cost $\top$ is called
{\em mandatory} (or {\em hard}). A clause with cost less than $\top$ is called
{\em non-mandatory} (or {\em soft}).

\begin{defn}
A Max-SAT instance is a pair $(\mathcal{F},\top)$ where $\top$ is a natural number and $\mathcal{F}$ is a set of weighted clauses with weights in the interval $[0..\top]$. The task of interest is to find an optimal model, if there is any.
\end{defn}

The following example shows that $\top$ can be used to express that we are only interested in assignments of a certain quality.

\begin{exmp}
Consider again the minimum vertex covering problem of the graph in Figure \ref{p_graph1} ($a$). With the new notation, the associated formula is

\[\begin{array}{ll}
          \mathcal{F} =  \{ & (\bar{x}_1,1), (\bar{x}_2,1), (\bar{x}_3,1), (\bar{x}_4,1), (\bar{x}_5,1), (x_1 \vee x_2,\top), (x_2 \vee x_3,\top), \\
          & (x_2 \vee x_4,\top),(x_2 \vee x_5,\top), (x_4 \vee x_5,\top)\}
            \end{array}
\]

which shows more clearly which clauses are truly weighted and which ones are mandatory. In the lack of additional information, $\top$ should be set to the sum of weights ($\top=5$), meaning that any assignment that satisfies the mandatory clauses should be taken into consideration.
Suppose now that somehow  (for example, with a local search algorithm) we find a covering of size $3$. We can set $\top$ to $3$ because any assignment with cost $3$ or higher does not interest us anymore. The resulting Max-SAT problem is tighter (and easier, because more partial assignments can be identified as unfeasible).
\end{exmp}

The interest of adding $\top$ to the problem formulation is twofold. On the one hand, it allows to explicit the mandatory nature of mandatory clauses. Besides, as we will see later, it allows to {\em discover} mandatory clauses that were {\em disguised} as weighted clauses. On the other hand, it allows to see SAT as a particular case of Max-SAT.

\begin{rem}  
A Max-SAT instance with $\top=1$ is essentially a SAT instance because there is no weight below $\top$. Consequently, every clause in the formula is mandatory.
\end{rem}

A weighted CNF formula may contain $(\Box,w)$ among its clauses. Since
$\Box$ cannot be satisfied, $w$ is a necessary cost of
any model. Therefore, $w$ is an explicit {\em lower bound}
of the cost of an optimal model. When the lower bound and the
upper bound have the same value (i.e., $(\Box, \top)\in \mathcal{F}$) the
formula is trivially unsatisfiable and we call this situation an {\em explicit
contradiction}. The idea of adding an upper bound $\top$ and a lower bound $(\Box,w)$ to the problem formulation was first proposed in the WCSP context \cite{Larrosa02}. 

\section{Extending SAT solving techniques to Max-SAT}

\subsection{Extending Simplification Rules and Clause Negation}

We say that two Max-SAT formulas are equivalent, $\mathcal{F}\equiv \mathcal{F}'$, if they contain the same set of variables, and complete assignments have the same costs. The following equivalence rules can be used to simplify CNF weighted formulas,
\begin{itemize}
\item {\em Aggregation}: $\{(A,w), (A,u)\} \equiv \{(A,w\oplus u)\}$
\item {\em Absorption}: $\{(A,\top), (A\vee B,w)\} \equiv \{(A,\top)\}$
\item {\em Unit clause reduction}: $\{(l,\top), (\bar{l}\vee A,w)\} \equiv \{(l,\top), (A,w)\}$
\item {\em Hardening}: If $\bigoplus_{i=1}^{k} u_i =\top$ and $\forall_{1\leq i<k} C_i\subset C_k$ then
$$\{(C_i,u_i)\}_{i=1}^{k-1}\cup\{(C_k,u_k)\}\equiv \{(C_i,u_i)\}_{i=1}^{k-1}\cup\{(C_k,\top)\}$$
%\item (Hardening I) if $u\oplus w=\top$ then $\{(\Box,u), (C,w)\} \equiv \{(\Box,u), (C,\top)\}$
\end{itemize}

{\em Aggregation} generalizes to Max-SAT the idempotency of the conjunction in classical SAT. 
{\em Absorption} rule indicates that in the Max-SAT context the absorbing clause must be mandatory. 
Similarly, {\em unit clause reduction} requires the unit clause being mandatory.
The correctness of these equivalences is direct and we omit the proof. The {\em Hardening} rule allows to identify weighted clauses that are indeed mandatory. It holds because the violation of $C_k$ implies the violation of all $C_{i}$ with $i<k$. Therefore, any assignment that violates $C_k$ will have cost $\bigoplus_{i=1}^k u_i=\top$.

It is easy to see that the {\em pure literal rule} can also be applied to Max-SAT. Besides, the assignment of a formula $\mathcal{F}[l]$ also holds in Max-SAT. As in SAT, it can be seen as the addition of $(l,\top)$ to the formula which allows a sequence of unit clause reductions followed by the application of the pure literal rule.

%Note that it allows to extend several SAT simplification rules that would be inapplicable with the standard definition of Max-SAT.

\begin{exmp}
Consider the following formula 
$\{(x,\top),(\bar{x},3),(y,8),(\bar{x}\vee\bar{y},3)\}$ with $\top=10$. We can apply unit clause reduction to the first and second clauses, which produces
$\{(x,\top),(\Box,3),(y,8),(\bar{x}\vee\bar{y},3)\}$. We can apply it again to the first and fourth clauses producing
$\{(x,\top),(\Box,3),(y,8),(\bar{y},3)\}$. The pure literal rule allows to remove the first clause producing
$\{(\Box,3),(y,8),(\bar{y},3)\}$.
We can harden the second clause because $3\oplus 8=\top$. Thus, we obtain 
$\{(\Box,3),(y,\top),(\bar{y},3)\}$. Unit clause reduction produces
$\{(\Box,3),(y,\top),(\Box,3)\}$. Aggregation yields $\{(\Box,6),(y,\top)\}$ and the pure literal rule produces the formula $\{(\Box,6)\}$ which trivially has an optimal model of cost $6$.
\end{exmp}

\begin{prop}
The algorithm that applies the previous simplifications until quiescence terminates in polynomial time.
\end{prop}

%\begin{pf}
%Termination follows the fact that all simplifications, either decrease the size of existing clauses, decrease the number of clauses in the formula, or increase the cost of an existing clauses. None of these things can be done infinitely many times.
%\end{pf}

Observe that if an explicit contradiction is achieved ({\em i.e.}, $(\Box,\top)\in \mathcal{F}$) all clauses are subsequently absorbed and the formula immediately collapses to $(\Box,\top)$.

The {\em negation of a weighted clause} $(C,w)$, noted $(\bar{C},w)$, means that the {\em satisfaction} of $C$ has cost $w$, while its negation is cost-free. Note that $\bar{C}$ is not clausal when $|C|>1$. In classical SAT the {\em De Morgan} rule can be used to recover the CNF syntax, but the following example shows that it cannot
be applied to weighted clauses.

\begin{exmp}
Consider the weighted clause $(x \vee y, 1)$ with $\top>1$. The truth table of its negation $(\overline{x \vee y},1)$ and the truth table of $\{(\bar{x},1), (\bar{y},1)\}$ are given below (ignore the last column for the moment). Note that they are not equivalent.
\begin{center}
\begin{tabular}{c|c|c|c}
$x\ y$ & $(\overline{x \vee y},1)$ & $\{(\bar{x},1), (\bar{y},1)\}$ & 
$\{(\bar{x}\vee \bar{y},1), (\bar{x} \vee y,1),(x \vee \bar{y},1) \}$\\
\hline
$\mathbf{f\ f}$ & $0$ & $0\oplus 0=0$ & $0\oplus 0\oplus 0=0$\\
$\mathbf{f\ t}$ & $1$ & $1\oplus 0=1$ & $0\oplus 0\oplus 1=1$\\
$\mathbf{t\ f}$ & $1$ & $0\oplus 1=1$ & $0\oplus 1\oplus 0=1$\\
$\mathbf{t\ t}$ & $1$ & $1\oplus 1=2$ & $1\oplus 0\oplus 0=1$\\
\end{tabular}
\end{center}

\end{exmp}

The following recursive transformation rule allows to recover the clausal form in totally or partially negated clauses. Let $A$ and $B$ be arbitrary disjunctions of clauses,
\[CNF(A\vee \overline{l\vee B},u) = \left\{ \begin{array}
            {c@{\quad \quad}l}
            (A\vee \bar{l},u) & :|B|=0\\
            \{(A\vee \bar{l} \vee B,u)\}\ \cup CNF(A\vee \bar{l} \vee \bar{B},u)\ \cup\\ 
            \cup\ CNF(A\vee l \vee \bar{B},u) &: |B|>0
            \end{array} \right.
\]

The last column in the truth table of the previous example shows the proper CNF encoding of clause 
$(\overline{x\vee y},1)$. The main drawback of this rule is that it generates an exponential number of new clauses
with respect the arity of the negated clause. 
We will show in Subsection 4.3 that it is possible to transform it into a linear number of 
clauses. 

\begin{thm}
  CNF($A\vee \overline{l\vee B},u$) returns an equivalent CNF expression.
\end{thm}
\begin{pf}
It is clear that CNF($A\vee \overline{l\vee B},u$) generates a CNF expression because the negation is applied to smaller sub-expression
at each recursive call. Eventually, it will be applied to literals, so the expression will be a clause.
We prove that CNF($A\vee \overline{l\vee B},u$) returns an equivalent expression by induction over $|B|$.
The $|B|=0$ is trivial since the left-hand and the right-hand sides are the same. Regarding the $|B|>0$ case, 
there are three ways to falsify $A\vee \overline{l\vee B}$. Each one of the three elements in the right-hand side corresponds to one of them. The last two are assumed correct by the induction hypothesis. 
\end{pf}

\begin{rem}
The weighted expression $(A\vee C \vee (\overline{C\vee B}),u)$, where $A$, $B$ and $C$ are disjunctions of literals, is equivalent to $(A\vee C \vee \bar{B},u)$, because they are falsified under the same circumstances.
\end{rem}

\subsection{Extending DPLL}
In Figure \ref{MaxDPLL} we present Max-DPLL, the
extension of DPLL to Max-SAT. Max-DPLL($\mathcal{F},\top$)
returns the cost of the optimal model if there is any, else it returns $\top$. 
First, the input formula is simplified with the rules from the previous subsection (line 1).
If the resulting formula is empty, there is a $0$ cost model (line 2).
If the resulting formula only contains the empty clause, the algorithm returns its 
cost (line 3).
Else, it selects a literal
$l$ (line 4) and makes two recursive calls (lines 5 and 6). In each call the formula is instantiated with $l$ and $\bar{l}$. 
Observe that the
first recursive call is made with the $\top$ inherited from its
parent, but the second call uses the output of the first call. This
implements the typical upper bound updating of depth-first branch and
bound. Finally, the best value of 
the two recursive calls is returned (line 7). Observe that, as search goes on, the value of $\top$ may decrease. Consequently, clauses that originally were soft may become hard which, in turn, may strengthen the potential of the simplification rules.
The parallelism with DPLL (Figure \ref{DPLL}) is obvious. The following statement shows that Max-DPLL is a true extension of classical DPLL.

\begin{rem}
The execution of Max-DPLL with a SAT instance (i.e., $(\mathcal{F},\top)$ with $\top =1$) behaves
like classical DPLL.
\end{rem}

It is easy to see that the time complexity of Max-DPLL is exponential on the number of variables $n$ and the space complexity is polynomial on $|\mathcal{F}|$. Therefore, DPLL and Max-DPLL have the same complexity.

\begin{figure}

\begin{tabbing}
mm\=mm\=mm\=mm\=mm\=mm\=mm\=mm\=\kill
\func{Max-DPLL}($\mathcal{F},\top$)\ \myret\ nat \\
1.\> $\mathcal{F} :=$ \call{Simplify}($\mathcal{F},\top$)\\
2.\> \myif\ $\mathcal{F}=\emptyset$ \ \mythen\ \myret\ $0$\\
3.\> \myif\ $\mathcal{F}=\{(\Box,w)\}$ \ \mythen\ \myret\ $w$\\
4.\> $l:=$\call{SelectLiteral}($\mathcal{F}$)\\
5.\> $\top:=$\call{Max-DPLL}($\mathcal{F}[l],\top$)\\
6.\> $\top:=$\call{Max-DPLL}($\mathcal{F}[\bar{l}],\top$)\\
7.\> \myret\ $\top$\\
\myendf\\
\end{tabbing}
\caption{If $(\mathcal{F},\top)$ has models, Max-DPLL
returns the optimal cost. Else it returns $\top$. \label{MaxDPLL}}
\end{figure}

\subsection{Extending the Resolution Rule}

Consider the {\em subtraction} of costs ($\ominus$) defined as in \cite{Schiex00}.
Let $u,w\in [0,\ldots,\top]$ be two weights such that $u\geq w$,
\[ u\ominus w = \left\{ \begin{array}
            {r@{\quad:\quad}l}
            u-w & u\neq \top\\
            \top & u = \top
            \end{array} \right.
\]

Essentially, $\ominus$ behaves like the usual subtraction except that $\top$ is an absorbing element. The resolution rule can be extended from SAT to Max-SAT as,

\[
\{(x \vee A, u),(\bar{x} \vee B, w)\} \equiv \left\{ \begin{array}
            {l@{\quad}}
            (A\vee B,m), \\
            (x\vee A,u\ominus m), \\
            (\bar{x}\vee B,w\ominus m), \\
            (x \vee A \vee \bar{B},m), \\
            (\bar{x} \vee \bar{A} \vee B,m) 
            \end{array} \right\}
\]
where $m=\min \{u,w\}$. 
In this rule, that we 
call Max-RES,
$(A\vee B,m)$ is called the {\em resolvent}; 
$(x\vee A,u\ominus m)$ and $(\bar{x}\vee B,w\ominus m)$ are called the {\em posterior clashing clauses}.
$(x \vee A \vee \bar{B},m)$ and $(\bar{x} \vee \bar{A} \vee B,m)$ are called the {\em compensation clauses}.
The effect of Max-RES, as in classical resolution, is to infer (namely, make explicit)
a connection between $A$ and $B$. 
However, there is an important difference between classical resolution and Max-RES. While classical resolution yields the {\em addition} of a new clause, Max-RES is a transformation rule. Namely, it requires the {\em replacement} of the left-hand clauses by the right-hand clauses. The reason is that some cost of the prior clashing clauses must be subtracted in order to {\em compensate} the new inferred information. Consequently, Max-RES is better understood as a {\em movement} of knowledge.

\begin{exmp}
If we apply Max-RES to the following clauses $\{(x\vee y, 3),(\bar{x}\vee y \vee z, 4)\}$ (with $\top>4$) we obtain 
$\{(y\vee y\vee z,3),
(x\vee y,3\ominus 3),
(\bar{x}\vee y \vee z, 4\ominus 3),
(x\vee y \vee \overline{(y\vee z)},3),
(\bar{x}\vee \bar{y}\vee y\vee z,3)\}$. The first and fourth clauses can be simplified. The second clause can be omitted because it weight is zero. The fifth clause can be omitted because it is a tautology. Therefore, we obtain the equivalent formula $\{(y\vee z,3),
(\bar{x}\vee y \vee z,1),
(x\vee y \vee \bar{z},3)\}$ 
\end{exmp}

The previous example showed that, under certain conditions, some of the right-hand side clauses can be removed.
Clause $(x\vee A,u\ominus m)$ (symmetrically for $(\bar{x}\vee B,w\ominus m)$) can be omitted iff either,\begin{itemize}
\item $B\subseteq A \wedge m=\top$, or
\item $u=m<\top$.
\end{itemize}
The first case holds because the clause is absorbed by the resolvent $(A,\top)$. The second case holds because $u\ominus m=0$.

Regarding clause $(x \vee A \vee \bar{B},m)$ (symmetrically for $(\bar{x} \vee \bar{A} \vee B,m)$), it can be omitted iff either,
\begin{itemize}
\item $B\subseteq A$, or
\item $u=\top$.
\end{itemize}
The first case holds because the clause is a tautology. The second case holds because the clause is absorbed by the posterior clashing clause $(x \vee A,\top\ominus m=\top)$.

\begin{rem}
The application of Max-RES to mandatory clauses is equivalent to classical resolution.
\end{rem}
\begin{pf}
Clashing clauses being mandatory means that $u=w=\top$. Clearly, $m=\min \{u,w\}=\top$, $u\ominus m=\top$ and $w\ominus m=\top$. Consequently, all right-hand clauses are mandatory. Therefore, the prior and posterior clashing clauses are equal. Overmore, the compensation clauses are absorbed by the clashing clauses (as we previously noted). Thus, Max-RES has the effect of  adding $(A\vee B, \top)$ to the formula, which is equivalent to classical resolution.
\end{pf}

\begin{thm}
Max-RES is sound. 
\end{thm}
\begin{pf}
The following table contains in the first columns all the truth
assignments, in the second column the cost of the assignment according
to the clauses on the left-hand of the Max-RES rule, and in the third
column the cost of the assignment according
to the clauses on the right-hand of the Max-RES rule. As it can be
observed, the costs the are same, so the resulting problem is equivalent.
\begin{center}
\begin{tabular}{c|c|c}
$x\ A\ B$ & Left & Right \\
\hline
$\mathbf{f\ f\ f}$ & $u$ & $m \oplus (u \ominus m)$\\
$\mathbf{f\ f\ t}$ & $u$ & $m \oplus (u \ominus m)$\\
$\mathbf{f\ t\ f}$ & $0$ & $0$\\
$\mathbf{f\ t\ t}$ & $0$ & $0$\\
$\mathbf{t\ f\ f}$ & $w$ & $m \oplus (w \ominus m)$\\
$\mathbf{t\ f\ t}$ & $0$ & $0$\\
$\mathbf{t\ t\ f}$ & $w$ & $m \oplus (w \ominus m)$\\
$\mathbf{t\ t\ t}$ & $0$ & $0$\\
\end{tabular}
\end{center}
\end{pf}

Observe that compensation clauses $(x \vee A \vee \bar{B},m)$ and $(\bar{x} \vee \bar{A} \vee B,m)$ are not in clausal form when $|A|>1$ and $|B|>1$. In the following, we assume that they are transformed to clausal form as needed.
In Subsection 4.1, we introduced a recursive rule that allows to recover the clausal form in totally or partially negated clauses. We noted that it produces an exponentially large number of new clauses. Interestingly, Max-RES allows to redefine it in such a way that only a linear number of clauses is generated,
\[CNF_{\mathtt{linear}}(A\vee \overline{l\vee B},u) = \left\{ \begin{array}
            {r@{\quad:\quad}l}
            A\vee \bar{l} & |B|=0\\
            \{(A\vee \bar{l} \vee B,u)\}\cup CNF_{\mathtt{linear}}(A \vee \bar{B},u)  & |B|>0
            \end{array} \right.
\]
The new rule is correct
because the two recursive calls of CNF (Subsection 4.1), $CNF(A\vee \bar{l} \vee \bar{B},u)$ and $CNF(A\vee l \vee \bar{B},u)$, can be resolved on literal $l$ and we obtain the equivalent call $CNF(A \vee \bar{B},u)$. For example, the application of $CNF_{\mathtt{linear}}$ to 
$(\overline{x \vee y},1)$ (Example 11) produces the equivalent $\{(\bar{x}\vee y,1),(\bar{y},1)\}$. Observe that the output of $CNF_{\mathtt{linear}}$ depends on how the literals are ordered in the clause.

\subsection{Extending DP}

\begin{figure}
\begin{tabbing}
mm\=mm\=mm\=mm\=mm\=mm\=mm\=mm\=\kill
\func{Max-VarElim}($\mathcal{F}, \top,x_i$)\ \myret\ weighted CNF formula \\
1.\> $\mathcal{B}:=\{(C,u)\in \mathcal{F}|\ x_i\in var(C)\}$\\
2.\> $\mathcal{F}:=\mathcal{F}-\mathcal{B}$\\
3.\> \mywhile\ $\exists (x_i\vee A,u) \in \mathcal{B}$\ \mydo\\
4\> \> $(x_i\vee A,u):=$\call{PopMinSizeClause}($\mathcal{B}$)\\
5.\> \> \mywhile\ $u>0 \wedge\ \exists_{(\bar{x}_i\vee B,w)\in \mathcal{B}}$ s.t. \call{Clash}($x_i\vee A,\bar{x}_i\vee B$)\ \mydo\\
6.\> \> \> $m:=\min \{u,w\}$\\
7.\> \> \>$u:=u\ominus m$\\
8.\> \> \>$\mathcal{B}:=\mathcal{B} - \{(\bar{x}_i\vee B,w)\} \cup \{(\bar{x}_i\vee B,w\ominus m)\}$\\
9.\> \> \>$\mathcal{B}:=\mathcal{B} \cup \{(x_i\vee A \vee \bar{B},m) \cup (\bar{x}_i\vee \bar{A} \vee B,m)\}$\\
10.\> \> \>$\mathcal{F}:=\mathcal{F} \cup \{(A\vee B,m)\}$\\
11.\> \>\myendw\\
12.\> \myendw\\
13. \> \myret\ ($\mathcal{F}$)\\
\myendf\\

\func{Max-DP}($\mathcal{F},\top$)\ \myret\ nat\\
14.\> $\mathcal{F} :=$ \call{Simplify}($\mathcal{F},\top$)\\
15.\> \myif\ $\mathcal{F}=\emptyset$ \ \mythen\ \myret\ $0$\\
16.\> \myif\ $\mathcal {F}=\{(\Box,u)\}$\ \mythen\ \myret\ $u$\\
17.\> $x_i:=$\call{SelectVar}($\mathcal{F}$)\\
18.\> \myret\ \call{Max-DP}(\call{VarElim}($\mathcal{F},\top,x_i$),$\top$)\\
\myendf\\
\end{tabbing}
\caption{If $(\mathcal{F},\top)$ has models, Max-DP returns their optimal cost. Else it returns $\top$. \label{MaxDP}}
\end{figure}

The following example shows that, unlike classical resolution, the unrestricted application of Max-RES does not guarantee termination\footnote{This fact was first observed in the WCSP context by \cite{Schiex00}}. 
\begin{exmp}
Consider the following formula $ \{(x\vee y,1),(\bar{x} \vee z,1)\}$ with $\top=3$.
If we apply Max-RES, we obtain
$ \{(y\vee z,1),(x \vee y \vee \bar{z},1),(\bar{x} \vee \bar{y} \vee z,1)\}$. If
we apply Max-RES to the first and second clauses we obtain
$ \{(x\vee y,1),(\bar{x} \vee y \vee z,1),(\bar{x} \vee \bar{y} \vee z,1)\}$. If we 
apply now Max-RES to the second and third clauses we obtain
$ \{(x\vee y,1),(\bar{x} \vee z,1)\}$, which is the initial formula.
\end{exmp}

Nevertheless, Bonet {\em et al.} \cite{Bonet06} have recently proved that when all clauses are non-mandatory, the directional application of Max-RES solves the Max-SAT problem. If their proof is combined with the proof of correctness of DP \cite{Davis60} (namely, all clauses being mandatory), we have that the extension of DP to Max-SAT produces a correct algorithm.
Max-DP (depicted in Figure \ref{MaxDP}) is the extension of DP to Max-SAT.
Both algorithms are essentially equivalent the main difference being that Max-DP performs Max-RES instead of classical resolution. 
Observe the parallelism between Function \texttt{VarElim} (Fig. \ref{DP}) and
Function \texttt{Max-VarElim} (Fig. \ref{MaxDP}). Both are in charge of the elimination of variable $x_i$ from the formula. As in the SAT case, \texttt{Max-VarElim} computes the bucket $\mathcal{B}$ (line 1) and removes its clauses from the formula (line 2). Then, it selects a clause $(x\vee A,u)$ and resolves it with all its clashing clauses. 
In \texttt{VarElim} clause $x\vee A$ is resolved until no clashing clauses exist.
In \texttt{Max-VarElim} clause $(x\vee A,u)$ is resolved until its weight $u$ decreases to $0$ or no clashing clauses exist. A worth noting difference with respect to the SAT case is that  \texttt{Max-VarElim} selects in line 4 a {\em minimal size} clause.
Such minor difference is not required for the correctness of the algorithm but only to achieve the complexity stated in Theorem 23.

The following lemma shows that \texttt{Max-VarElim} transforms the input formula preserving  its optimality.
\begin{lem}
Consider a call to the \texttt{Max-VarElim} function. Let $(\mathcal{F},\top)$ denote the input formula and let $(\mathcal{F}',\top)$ denote the output formula. It is true that $(\mathcal{F},\top)$  has models iff $(\mathcal{F}',\top)$ has models. Besides, if $(\mathcal{F},\top)$ has models, the cost of the optimal one is the same as the cost of the optimal model of $(\mathcal{F}',\top)$.
\end{lem}
\begin{pf}
See Appendix A.
\end{pf}

\begin{thm}
Algorithm Max-DP is correct.
\end{thm}
\begin{pf}
Max-DP is a sequence of variable eliminations until variable-free formula is obtained. Lemma 19 shows that each transformation preserves the cost of the optimal model. Therefore, the cost of the final variable-free formula $(\Box,u)$ is the cost of the optimal model of the original formula.
\end{pf}

The following lemma, shows that it has the same complexity to eliminate a variable in classical SAT and in Max-SAT.
\begin{lem}
Let $(\mathcal{F}, \top)$ be a Max-SAT instance and $x_i$ one of its variables.
Let $n_i$ denote the number of variables sharing some clause with $x_i$ in $\mathcal{F}$.
The space and time complexity of \texttt{Max-VarElim}($\mathcal{F},\top,x_i$) is $O(3^{n_i})$ and $O(9^{n_i})$, respectively.
\end{lem}
\begin{pf}
See Appendix A.
\end{pf}

The next lemma, shows that the induced graph plays the same role in DP and in Max-DP.
\begin{lem}
Let $d$ denote the reverse order in which Max-DP($\mathcal{F},\top$) eliminates variables.
The width of $x_i$ along $d$ in the induced graph $G(\mathcal{F})^*_d$ bounds above the number of variables sharing some clause with $x_i$ at the time of its elimination.
\end{lem}
\begin{pf}
Same as the SAT case (Lemma 3).
\end{pf}

The following theorem, which trivially follows from the previous two lemmas, bounds the complexity of Max-DP.
\begin{thm}
Let $(\mathcal{F}, \top)$ be an arbitrary Max-SAT instance.
Let $d$ denote the reverse order in which Max-DP($\mathcal{F}, \top$) eliminates variables. 
The space and time complexity of DP($\mathcal{F}$) is $O(n\times 3^{w^*_d})$ and $O(n\times 9^{w^*_d})$, respectively, where $w^*_d$ is the induced width of the interaction graph $G(\mathcal{F})$ along $d$.
\end{thm}

Observe that the complexities of DP and Max-DP are the same, even though Max-SAT has a complexity higher than SAT. The same phenomenon has already been observed with respect to CSP and its optimization version WCSP when using the {\em bucket-elimination} \cite{Dechter99} algorithm. Note that bucket-elimination is a meta-algorithm based on the variable-elimination principle and DP and Max-DP are particular instantiations of it. The following remark shows that Max-DP is a true extension of DP.

\begin{rem}
The execution of Max-DP with a SAT instance (i.e., $(\mathcal{F},\top)$ with $\top =1$) behaves
like classical DP.
\end{rem}

\section{Efficient Inference}

The complexity results  of the previous section show that solving Max-SAT with pure resolution methods is 
in general too space consuming and can only be used in practice with formulas with a small induced width (around $30$ with current computers). A natural alternative is to use only restricted forms of resolution that simplify the formula and use search afterwards. 
In this Section we summarize some simplification rules that have been proposed in the recent Max-SAT literature and show that they can be naturally explained with our framework. We also introduce two original ones that will be used in the solver that we will introduced in Section 6.

We classify these simplification rules in three categories: single applications of resolution, multiple applications of resolution (namely, hyper-resolution), and variable elimination.

\subsection{Single Resolution}

\begin{prop}
Unit clause reduction (also called {\em upper bound rule} in \cite{Zhang05}),
$$\{(l,\top), (\bar{l}\vee A,w)\} \equiv \{(l,\top), (A,w)\}$$
is a particular case of Max-RES.
\end{prop}
\begin{pf}
If $w=\top$, we have the classical SAT case, which is trivial. If $w<\top$,
we have that the application of Max-RES to $\{(l\vee \Box,\top),(\bar{l}\vee A,w)\}$ produces 
$\{(A,w),
(l,\top \ominus \top),
(\bar{l}\vee A,w\ominus w), 
(l \vee \Box \vee \bar{A},w),
(\bar{l} \vee \neg\Box \vee A,u)\}$

The third clause can be removed because $w\ominus w=0$.
The fourth clause can be removed because it is absorbed by the second.
The fifth clause can be removed because it is a tautology.
\end{pf}

\shrink{
As a matter of fact, unit clause reduction can be generalized as,
$$\{(l \vee B, \top),(\bar{l} \vee A \vee B, w)\} \equiv \{(A\vee B,w), (l\vee B,\top)\}$$
which we call {\em subset resolution} because it requires the hard clause to be a proper subset of the other (except for the clashing variable). Subset resolution  has the nice effect of removing a literal from a clause.
}

\begin{prop}
{\em Neighborhood resolution} \cite{Larrosa05a} (also called {\em replacement of almost common clauses} in \cite{Bansal99}),
$$\{(l \vee A, u),(\bar{l} \vee A, w)\} \equiv \{ (A,w),(l\vee A,u\ominus w)\} $$
where, without loss of generality, $w\leq u$, is a particular case of Max-RES.
\end{prop}
\begin{pf}
Resolving the two left-hand clauses, we obtain $\{(A,w),
(l\vee A,u\ominus w),
(\bar{l}\vee A,w\ominus w), 
(l \vee A \vee \bar{A},w), 
(\bar{l} \vee A \vee \bar{A},w)\}$.
The third clause can be omitted because either its weight is $0$ (when $w<\top$), or it is absorbed by the resolvent (when $w=\top$). The fourth and fifth clauses can be omitted because they are tautologies.
\end{pf}

The simplification potential of neighborhood resolution is shown in the following example,
\begin{exmp}
Consider the formula 
$\{(z \vee y,1),(\bar{y}\vee z,1), (\bar{z},1)\}$. The application of neighborhood resolution yields $\{(z,1), (\bar{z},1)\}$ which allows a new application of neighborhood resolution producing the trivial formula $\{(\Box,1)\}$
\end{exmp}

The term {\em neighborhood resolution} was coined by \cite{Cha96} in the SAT context. The Max-SAT extension was first proposed in \cite{Bansal99}. The practical efficiency of the $|A|=0,1,2$ cases was assessed in \cite{Freuder96,Alsinet03}, \cite{Alsinet05} and \cite{Larrosa05a}, respectively. In the WCSP context, it is related to the notion of {\em projection} and has been used to enforce {\em node} and {\em arc-consistency} \cite{Schiex00,Larrosa02}.

\subsection{Variable elimination}

\begin{prop}
The {\em pure literal rule} (first proposed in the Max-SAT context in \cite{Bansal99}) is a special case of \texttt{Max-VarElim}
\end{prop}
\begin{pf}
Consider a formula $\mathcal{F}$ such that there is a literal $l$, whose negation does not appear in the formula. Let $x=var(l)$.
Function \texttt{Max-VarElim}$(\mathcal{F},\top,x)$ has the same effect as the pure literal rule, because there is no pair of clauses clashing on $x$. Thus, no resolution will be performed and all clauses containing  $l$ will be removed from the formula.
\end{pf}

\begin{prop}
The {\em elimination} rule \cite{Bansal99} (also called {\em resolution} in \cite{Niedermeier00,Chen2002}) which says that
if  $\mathcal{F}=\{(l\vee A,u),(\bar{l}\vee B,w)\}\cup \mathcal{F}'$ and $var(l)$ does not
occur in $\mathcal{F}'$ then
$$\mathcal{F}\equiv \mathcal{F}'\cup \{(A\vee B, \min\{u,w\})\}$$
is a special case of \texttt{Max-VarElim}
\end{prop}
\begin{pf} 
Let $x$ be the clashing variable (namely, $x=var(l)$). We need to prove that
Function \texttt{Max-VarElim} with $x$ as the elimination variable replaces $\{(l\vee A,u),(\bar{l}\vee B,w)\}$ by $\{(A\vee B, \min\{u,w\})\}$. There are two possibilities: If $\{(l\vee A,u),(\bar{l}\vee B,w)\}$ clash, they will be resolved and $(A\vee B, \min\{u,w\})$ will be added to the formula. All the clauses in the bucket after the resolution step do not clash on $x$, so \texttt{Max-VarElim} will discard them. If $\{(l\vee A,u),(\bar{l}\vee B,w)\}$ do not clash, \texttt{Max-VarElim} will directly discard them. In that case, $A\vee B$ either is a tautology or is absorbed, so it has no effect on the right-hand side of the elimination rule.
\end{pf}

\begin{prop}
Let $\tilde{x}$ denote either $x$ or $\bar{x}$. The {\em small subformula rule} \cite{Niedermeier00}, which says that, if  $\mathcal{F}=\{(\tilde{x}\vee \tilde{y}\vee A,u), (\tilde{x}\vee \tilde{y}\vee B,w), (\tilde{x}\vee \tilde{y}\vee C,v)\}\cup \mathcal{F}'$ and $x, y$ do not occur in $\mathcal{F}'$ 
then  
$$\mathcal{F}\equiv \mathcal{F}'$$
is a special case of \texttt{Max-VarElim}
\end{prop}
\begin{pf} We only need to prove that if we eliminate $x$ and $y$ from $\{(\tilde{x}\vee \tilde{y}\vee A,u), (\tilde{x}\vee \tilde{y}\vee B,w), (\tilde{x}\vee \tilde{y}\vee C,v)\}$ with function \texttt{Max-VarElim}, we obtain the empty formula $\varnothing$.

If all the occurrences of $x$ or $y$ have the same sign, the rule holds because to the pure literal rule can be applied. If there are occurrences of different sign, there are only two cases to consider (all other cases are symmetric):
\begin{itemize}
\item If we have $\{(x\vee y\vee A,u), (x\vee y\vee B,v), (\bar{x}\vee \bar{y}\vee C,w)\}$, there are no clauses clashing on $x$ (neither on $y$), so \texttt{Max-VarElim} will just discard the clauses.
\item If we have $\{(x\vee y\vee A,u),(\bar{x}\vee y\vee B,v),(x\vee \bar{y}\vee C,w)\}$, the first and second clauses clash, so Max-RES produces,
$$\{(y\vee A\vee B,m),
(x\vee y\vee A,u\ominus m),(\bar{x}\vee y\vee B,v\ominus m),
(x\vee y\vee A \vee \overline{y\vee B},m),$$
$$(\bar{x}\vee \overline{y\vee A} \vee y\vee C,m),
(x\vee \bar{y}\vee B,w)
\}$$
which is equivalent to,
$$\{(y\vee A\vee B,m),
(x\vee y\vee A,u\ominus m),(\bar{x}\vee y\vee B,v\ominus m),
(x\vee y\vee A \vee \bar{B},m),(\bar{x}\vee \bar{A} \vee y\vee C,m),$$
$$(x\vee \bar{y}\vee B,w)
\}$$
There are no further clauses clashing on $x$, so \texttt{Max-VarElim} will just discard all the clauses that mention it, producing the equivalent $\{(y\vee A\vee B,m)\}$. The pure literal rule will eliminate the clause, producing the empty formula.
\end{itemize}
\end{pf}

\subsection{Hyper-resolution}

\begin{figure}[t]
        \centering
	\fbox{\includegraphics*[4cm,21cm][12.7cm,26cm]{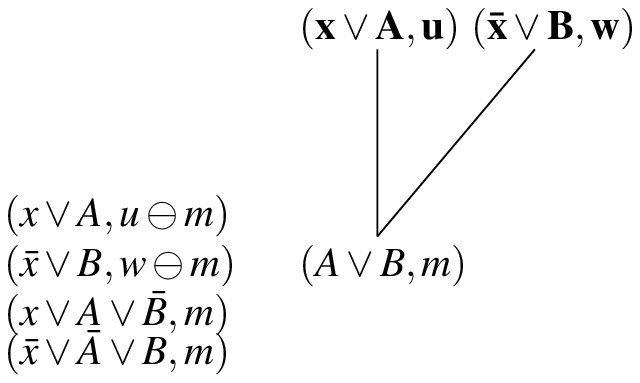}}
	\caption{Graphical representation of Max-RES.}
\label{d_res}
\end{figure}

{\em Hyper-resolution} is a well known SAT concept that refers to the compression of several resolution steps into one single step. In the following, we introduce four hyper-resolution inference rules. The first two ({\em star rule} and {\em dominating unit-clause}) are formal descriptions of already published rules. The other two rules ({\em cycle} and {\em chain resolution}) are original. We prove the correctness of these rules by developing the resolution tree that allows to transform the left-hand side of the rule into the right-hand side. Figure \ref{d_res} shows the graphical representation of Max-RES. On top there are the two prior clashing clauses. We write them in bold face to emphasize that they are removed from the formula. The resolvent is linked to the prior clashing clauses. At the left of the resolvent, we write the posterior clashing clauses and the compensation clashing clauses, which must be added to preserve equivalence.

\subsubsection{Star rule}

\begin{figure}[t]
        \centering
	\fbox{\includegraphics*[4cm,11.5cm][14.7cm,26cm]{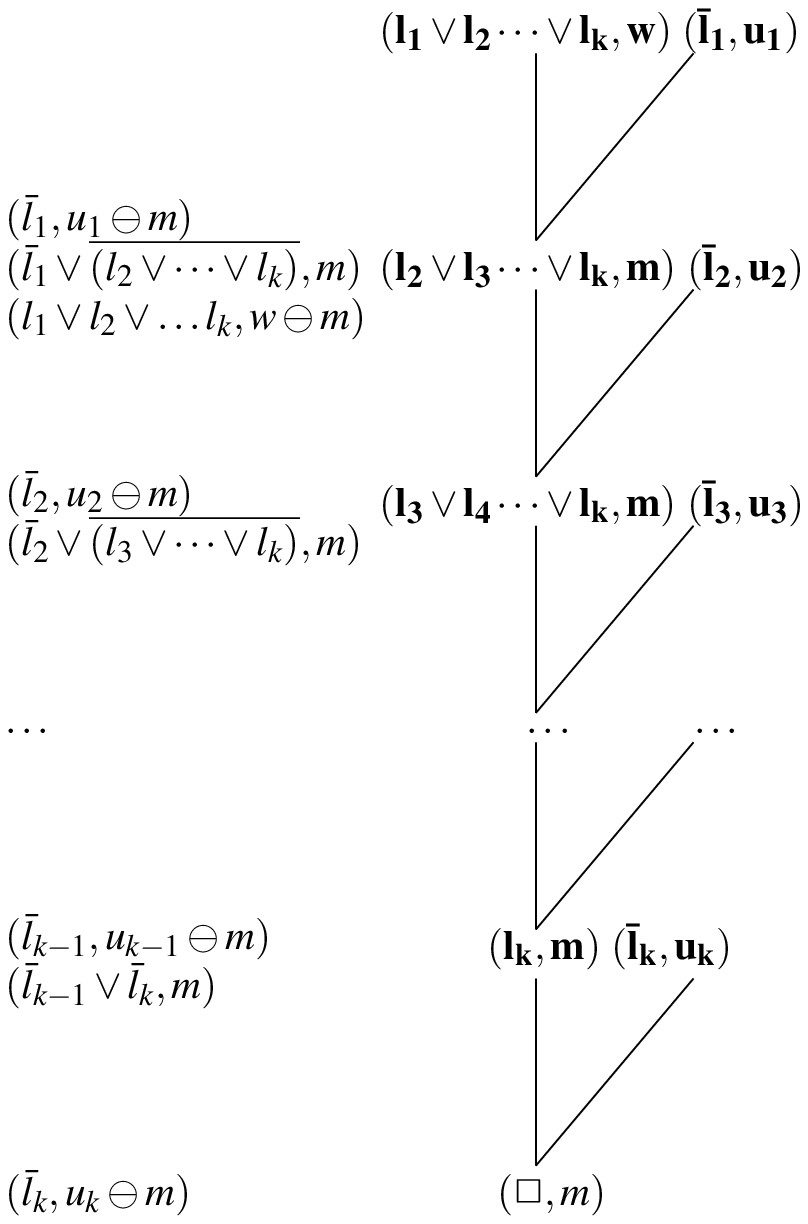}}
	\caption{Resolution tree of the {\em star rule}.}
\label{d_star}
\end{figure}

The {\em star rule} \cite{Niedermeier00,Alsinet05} identifies a clause of length $k$ such that each of its literals appears negated in a unit clause. Then, at least one of the clauses will be violated. Formally,
\[
\left\{ \begin{array}
            {l@{\quad}}
            (l_1\vee l_2 \vee \ldots l_k,w), \\
            (\bar{l}_i,u_i)_{1\leq i\leq k},\\
            \end{array} \right\}
\equiv 
\left\{ \begin{array}
            {l@{\quad}}
            (l_1\vee l_2 \vee \ldots l_k,w\ominus m), \\
            (\bar{l}_i\vee \overline{(l_{i+1}\vee l_{i+2}\vee \ldots \vee l_k)},m)_{1\leq i < k},\\
            (\bar{l}_i,u_i\ominus m)_{1\leq i\leq k},\\
            (\Box, m)\\
            \end{array} \right\}
\]
where $m=\min \{w,u_1,u_2,\ldots,u_k\}$.

This rule can be proved in $k$ resolution steps. Assume, without loss of generality that $\forall_{1\leq i<k} \ \ u_i\leq u_{i+1}$. 
Assume as well that $u_k<\top$ (otherwise unit clause reduction could have been previously triggered). Let $m=\min \{w,u_1\}$. Figure \ref{d_star} shows the corresponding resolution tree. 
Recall that bold clauses are resolved, so they must be removed from the formula. Essentially, each unit clause is used to eliminate one literal from the length $k$ clause. At the end, we derive the empty clause.

\subsubsection{Dominating unit-clause}

\begin{figure}[t]
        \centering
	\fbox{\includegraphics*[4cm,11.5cm][16.7cm,26cm]{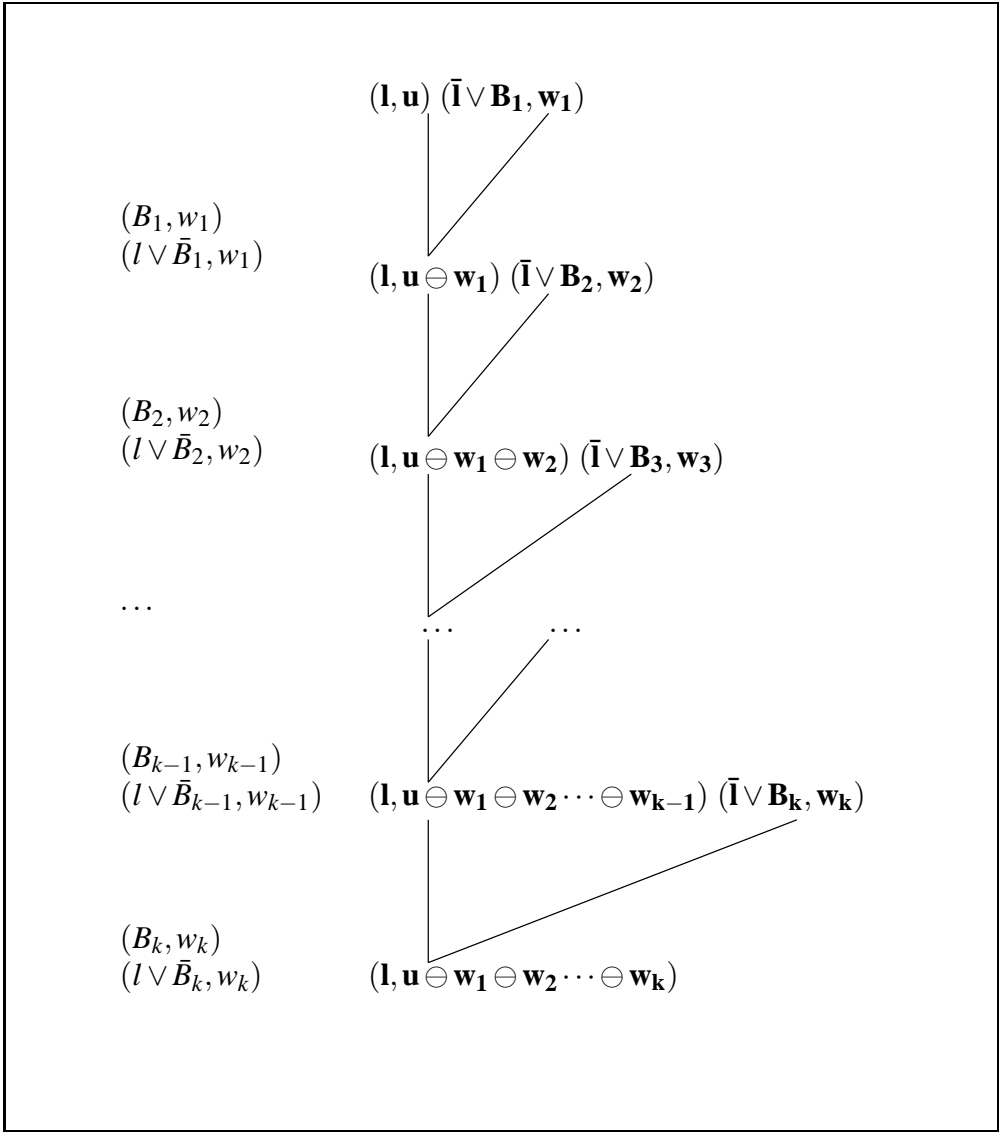}}
	\caption{Resolution tree the {\em dominating unit clause rule}.}
\label{d_duc}
\end{figure}

The {\em dominating unit-clause} rule \cite{Niedermeier00} (also called UP3 in \cite{Zhang05}) says that if the weight if a unit clause $(l,u)$ is higher than the sum of weights in which $\bar{l}$ appears, we can safely assign $\bar{l}$ to the formula. Formally,
$$\mathcal{F}=\{(l,u)\} \cup \{(l\vee A_i,u_i)\}_{i=1}^{k'} \cup \{(\bar{l}\vee B_j,w_j)\}_{j=1}^{k} \cup \mathcal{F}'$$ 
with $u\geq \sum_{j=1}^k w_j$ and $\mathcal{F}'$ does not contain any occurrence of $l$ or $\bar{l}$, then
$$\mathcal{F}\equiv \{(B_j,w_j)\}_{j=1}^{k} \cup \mathcal{F}'$$
This rule can be proved in $k$ resolution steps plus the application of the pure literal rule. Figure \ref{d_duc} shows the corresponding resolution tree. As in the previous case, we can assume that weight $u$ is less than $\top$ because otherwise the unit clause reduction could have been triggered.
At each step unit clause $l$ is resolved with one $(\bar{l}\vee B_j,w_j)$. Since, by definition, the weight of $l$ is larger than or equal to $w_j$, clause $\bar{l}\vee B_j$ is replaced by $B_j$. At the end of the process, there is no clause mentioning $\bar{l}$, so the pure literal rule can be applied, which proves the correctness of the rule.

\subsubsection{Chain resolution}

\begin{figure}[t]
        \centering
	\fbox{\includegraphics*[4cm,11.5cm][14.7cm,26cm]{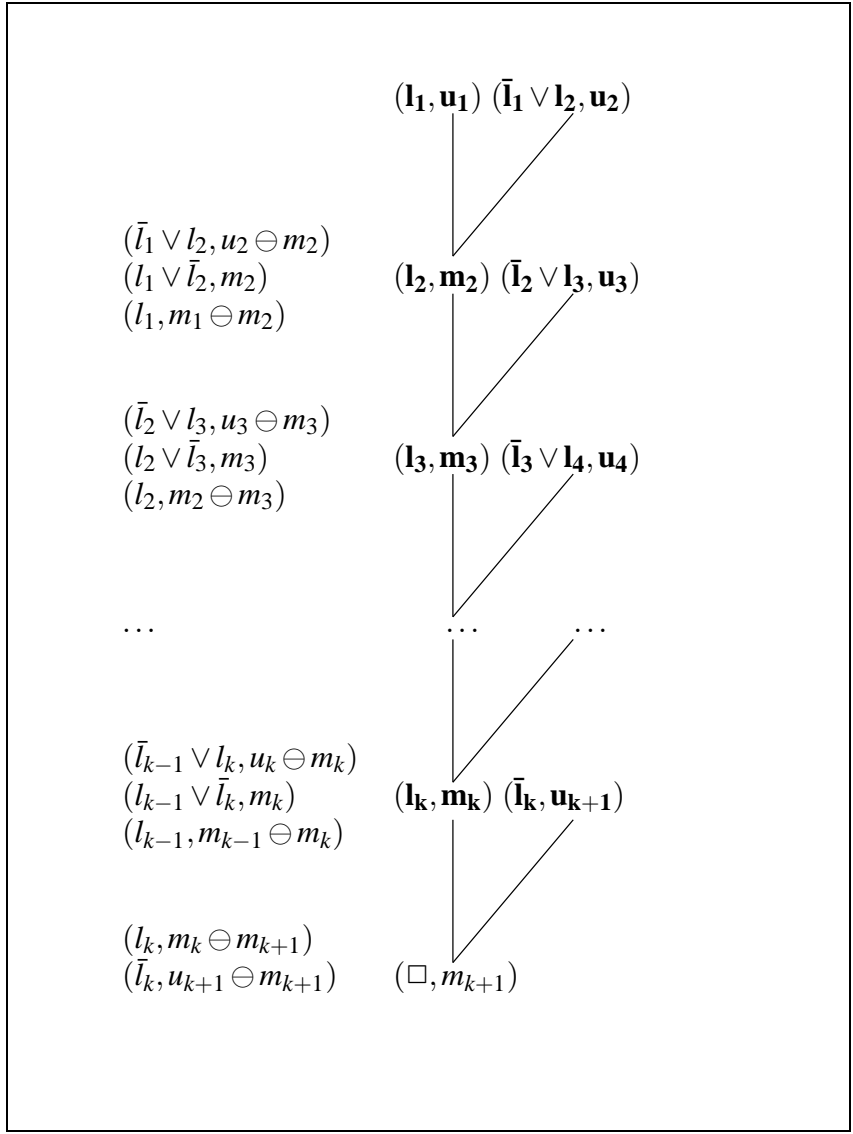}}
	\caption{Resolution tree of {\em chain resolution}.}
\label{d_chain}
\end{figure}

Our original {\em chain resolution} rule, identifies a subset of {\em chained} binary clauses and two unit clauses associated to the ends. When such pattern exists, a sequence of unit resolution steps suffices to derive the empty clause. The rule is the following,

\[
\left\{ \begin{array}
            {l@{\quad}}
            (l_1,u_1), \\
            (\bar{l}_i\vee l_{i+1},u_{i+1})_{1\leq i <k},\\
            (\bar{l}_k,u_{k+1})\\
            \end{array} \right\}
\equiv 
\left\{ \begin{array}
            {l@{\quad}}
            (l_i,m_i\ominus m_{i+1})_{1\leq i \leq k}, \\
            (\bar{l}_i\vee l_{i+1},u_{i+1}\ominus m_{i+1})_{1\leq i <k},\\
            (l_i\vee \bar{l}_{i+1},m_{i+1})_{1\leq i <k},\\
            (\bar{l}_k,u_{k+1}\ominus m_{k+1})\\
            (\Box, m_{k+1})\\
            \end{array} \right\}
\]
where $m_i=\min \{u_1,u_2,\ldots,u_i\}$ and $\forall_{1\leq i<j \leq k}\ \ var(l_i)\neq var(l_j)$.
This rule can also be proved in $k$ steps of resolution. Figure \ref{d_chain} shows the corresponding resolution tree. Starting with unit clause $l_1$, at each resolution step a unit clause $l_i$ is resolved with $(\bar{l}_i\vee l_{i+1},u_{i+1})$, which produces the unit clause $l_{i+1}$ to be used in the following resolution step. The last unit clause obtained is $l_{k}$ and it is resolved with $(\bar{l}_k,u_{k+1})$, which derives the empty clause.

\begin{exmp}
Consider the following formula $\{(x,2),(\bar{x}\vee y,1),(\bar{y}\vee z,\top),(\bar{z},2)\}$. If we resolve $(x,2)$ and $(\bar{x}\vee y,1)$ we obtain 
$\{(x,1),(y,1),(x\vee \bar{y},1),(\bar{y}\vee z,\top),(\bar{z},2)\}$. If we resolve $(y,1)$ and $(\bar{y}\vee z,\top)$ we obtain 
$\{(x,1),(x\vee \bar{y},1),(z,1),(y\vee \bar{z},1),(\bar{y}\vee z,\top),(\bar{z},2)\}$. Next, if we resolve $(z,1)$ and $(\bar{z},2)$, we obtain 
$\{(x,1),(x\vee \bar{y},1),(y\vee \bar{z},1),(\bar{y}\vee z,\top),(\bar{z},1),(\Box,1)\}$
\end{exmp}

Observe that chain resolution with $k=1$ reduces to neighborhood resolution, with $k=2$ reduces to the {\em star rule}, with $k=3$, it is the 3-RES rule proposed in \cite{Larrosa06}. Chain resolution with $k=2$ is also related to the enforcement of {\em existential arc consistency} in WCSP \cite{deGivry05}.

\subsubsection{Cycle Resolution}

\begin{figure}[t]
        \centering
	\fbox{\includegraphics*[4cm,10.5cm][14.7cm,25cm]{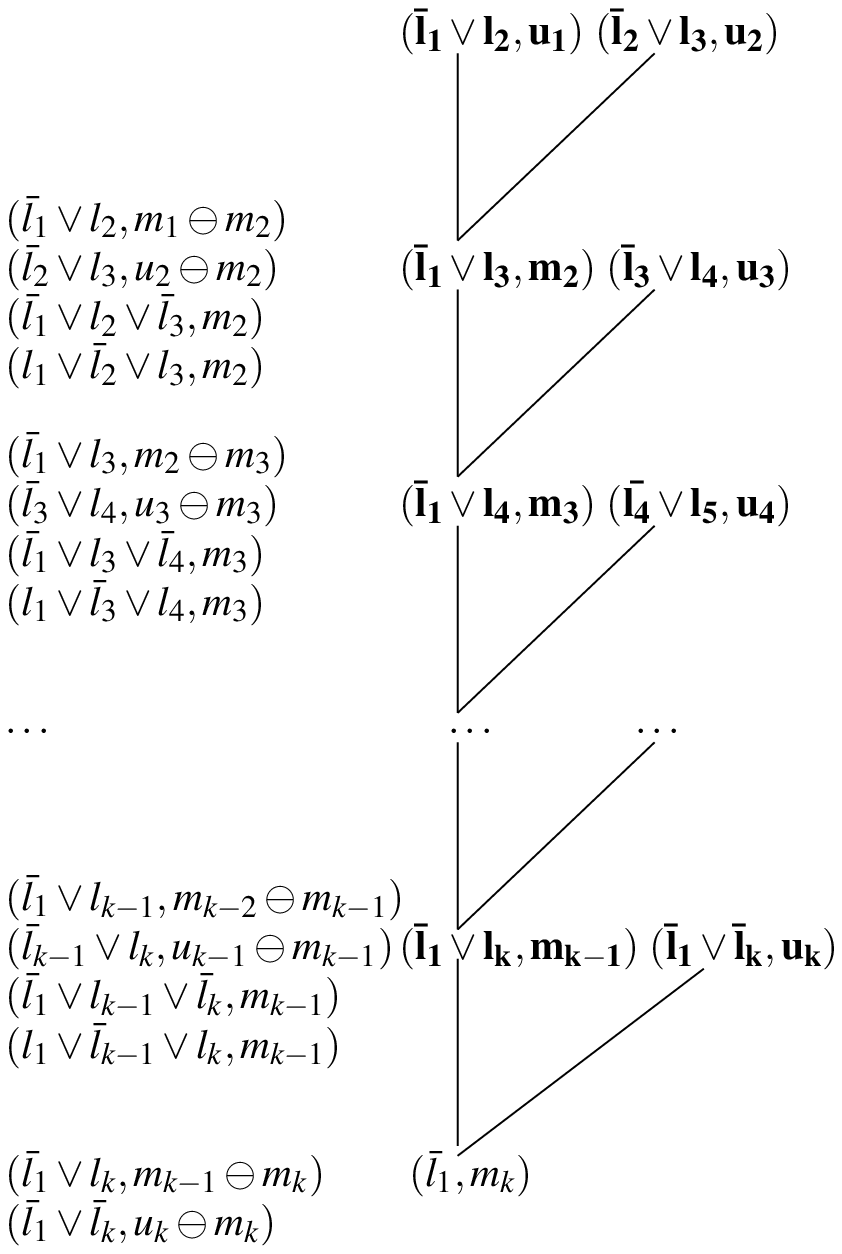}}	
	\caption{Resolution tree of {\em cycle resolution}.}
\label{d_cycle}
\end{figure}

Our original {\em cycle resolution}, identifies a subset of binary clauses with a cyclic structure. When such a pattern exists, a sequence of resolution steps with binary clauses suffices to derive a new unit clause. The rule is the following,

\[
\left\{ \begin{array}
            {l@{\quad}}
            (\bar{l}_i\vee l_{i+1},u_{i})_{1\leq i <k},\\
            (\bar{l}_1 \vee \bar{l}_k,u_{k})\\
            \end{array} \right\}
\equiv 
\left\{ \begin{array}
            {l@{\quad}}
(\bar{l}_1\vee l_{i},m_{i-1}\ominus m_{i})_{2\leq i \leq k},\\
(\bar{l}_i\vee l_{i+1},u_{i}\ominus m_{i})_{2\leq i <k},\\
(\bar{l}_1\vee l_i \vee \bar{l}_{i+1},m_i)_{2\leq i <k},\\
(l_1\vee \bar{l}_i \vee l_{i+1},m_i)_{2\leq i <k},\\
(\bar{l}_1\vee \bar{l}_{k}, u_k \ominus m_k),\\
(\bar{l}_1,m_{k})
           \end{array} \right\}
\]
where $m_i=\min \{u_1,u_2,\ldots,u_i\}$ and $\forall_{1\leq i<j \leq k}\ \ var(l_i)\neq var(l_j)$. This rule can be proved in $k-1$ steps of resolution. Figure \ref{d_cycle} shows the corresponding resolution tree.
The use of the cycle rule is to derive new unit clauses that, in turn, can be used by chain resolution to increase the weight of the empty clause.

\begin{exmp}
Consider the formula $\{(x_1\vee x_2,1), (\bar{x}_1\vee x_3,1), (\bar{x}_2\vee x_3,1), (\bar{x}_3\vee \bar{x}_4,1), (x_4\vee x_5,1), (\bar{x}_5,1)\}$. We can apply the cycle rule to the three first clauses obtaining,
$\{(x_3,1), (x_1\vee x_2\vee \bar{x}_3,1), (\bar{x}_1\vee \bar{x}_2\vee x_3,1),
(\bar{x}_3\vee \bar{x}_4,1), (x_4\vee x_5,1), (\bar{x}_5,1)\}$.
Chain resolution can be applied to the unary and binary clauses producing,
$\{(x_1\vee x_2\vee \bar{x}_3,1), (\bar{x}_1\vee \bar{x}_2\vee x_3,1),
(x_3\vee x_4,1), (\bar{x}_4\vee \bar{x}_5,1), (\Box,1)\}$.
\end{exmp}

Observe that cycle resolution with $k=3$ is one particular case of the so-called {\em high-order} consistencies proposed in \cite{Cooper05} for WCSP. In particular, it is a weighted version restricted to boolean variables of {\em path inverse consistency}  \cite{Freuder96}.

\section{An efficient Max-SAT solver}

\begin{figure}[t]
\begin{tabbing}
mm\=mm\=mm\=mm\=mm\=mm\=mm\=mm\=\kill
\func{Simplify}($\mathcal{F},\top$) \\
1.\> $stop := false$ \\
2.\> \mydo\ \\
3.\>  \>  \myif\ $(l,\top) \in \mathcal{F}$ \mythen\ \myapply\  $\mathcal{F}[l]$\\
4.\>  \>  \myelseif\ $\{(C,u), (C,w)\} \subseteq \mathcal{F}$\mythen\ \myapply\ \texttt{Aggregation}\\
5.\>  \>  \myelseif\ $\{(\Box,u), (C,w)\} \subseteq \mathcal{F} \land u\oplus w=\top$\ \mythen\ \myapply\  \texttt{Hardening}\\
6.\>  \>  \myelseif\ $\{(x\vee A,u), (\bar{x}\vee A,w)\} \subseteq \mathcal{F}$ \mythen\ \myapply\ \texttt{Neighbourhod Res.} \\
7.\>  \>  \myelseif\ $\{(l_1,u_1), (\bar{l}_i\vee l_{i+1},u_{i+1})_{1\leq i <k},(\bar{l}_k,u_{k+1})\} \subseteq \mathcal{F}$ \mythen\ \myapply\ \texttt{Chain Res.}\\
8.\>  \>  \myelseif\ $\{\{(l \vee h,u),(\bar{l}\vee q,v),(\bar{h}\vee q,w)\}\}\subseteq \mathcal{F}$ \mythen\ \myapply\ \texttt{Cycle Res.} \\
9.\> \myuntil\ $(((\Box,\top) \in \mathcal{F}) \vee stop)$\\
10. \> \myret\ ($\mathcal{F}$)\\
\myendf\\

\func{Max-DPLL}($\mathcal{F},\top$)\ \myret\ nat \\
11.\> $\mathcal{F} :=$ \call{Simplify}($\mathcal{F},\top$)\\
12.\> \myif\ $\mathcal{F}=\emptyset$ \ \mythen\ \myret\ $0$\\
13.\> \myif\ $\mathcal{F}=\{(\Box,w)\}$ \ \mythen\ \myret\ $w$\\
14.\> $l:=$\call{SelectLiteral}($\mathcal{F}$)\\
15.\> $\top:=$\call{Max-DPLL}($\mathcal{F}[l],\top$)\\
16.\> $\top:=$\call{Max-DPLL}($\mathcal{F}[\bar{l}],\top$)\\
17.\> \myret\ $\top$\\
\myendf\\
\end{tabbing}
\caption{\texttt{Max-DPLL} enhanced with inference. Function \texttt{Simplify}$(\mathcal{F},\top)$ converts the input formula into a simpler one. Note that in our implementation, for efficiency reasons, we only consider the $|A|\leq 1$ and $|C|\leq 2$ case.}
\label{f_dpll}
\end{figure}

In the previous section we presented a set of simplification rules. Some of them have been previously proposed by other researchers, while some others are original. We showed that all of them can be view as special cases of resolution, hyper-resolution or variable elimination. In this Section we consider their incorporation into the Max-DPLL algorithm introduced in Subsection 4.2. The idea is to use these rules to simplify the current Max-SAT formula before letting Max-DPLL branch on one of the variables. Our experimental work indicates that it is not cost effective to apply all of them on a general basis. We have observe that only three rules are useful in general: {\em neighborhood resolution}, {\em chain resolution} and {\em cycle resolution}. Besides, it only pays off to apply these rules to clauses of very small size (up to 2). The reason being that there is only a quadratic number of them which bounds the overhead of the detection of situations when they can be triggered. 
Regarding cycle resolution, we only found effective to apply the $k=3$ case (namely, considering triplets of variables).
Note that the fact that our solver only incorporates these three rules, does not prevent other rules from being effective in classes of problems where we did not experiment. 

A high-level description of our solver appear in Figure \ref{f_dpll}. It is Max-DPLL augmented with the simplification rules in function \texttt{Simplify}. This function iteratively simplifies the formula. It stops when a contradiction is derived or no further simplification can be done (line 9).
Simplification rules are arranged in an ordered manner, which means that if two rules $R$ and $R'$ can be applied, and rule $R$ has higher priority than rule $R'$, the algorithm will chose $R$.
The rules with the highest priority are {\em unit clause reduction} and {\em absorption} grouped in the assignment $\mathcal{F}[l]$ operation (line 3). Next, we have {\em aggregation} (line 4), {\em hardening} (line 5), {\em neighborhood resolution} (line 6), {\em chain resolution} (line 7) and
{\em cycle resolution} restricted to cycles of length $3$ (line 8).

Although our actual implementation is conceptually equivalent to the
pseudo-code of Figure \ref{f_dpll} it should be noted that such code aims at clarity and 
simplicity. Thus, a direct translation into a programming language is highly inefficient. The main source of inefficiency 
is the time spent searching for clauses that match with the left-hand side of the simplification rules. 
This overhead, which depends on the number of clauses, takes place at each iteration of the loop.
As we mentioned, our current implementation only takes into account clauses of arity less than or equal to two. Another way to decrease such overhead is to identify those events that may raise the applicability of the transformations. For instance, a clause may be made mandatory (line 5) only when its weight or the weight of the empty clause increases. Then, our implementation reacts to these events and triggers the corresponding rules.
Such approach is well-known in the constraint satisfaction field and it is usually implemented with streams of pending events \cite{Bessiere94,deGivry05}.

The way in which we detect the chain resolution pattern also deserves special consideration. At each search node, we consider the set of binary and unary clauses and compute the corresponding implication graph defined as follows: 
\begin{itemize}
\item for each variable $x_i$, the graph has two vertices $x_i$ and $\bar{x}_i$,
\item for each binary clause $(l_i\vee l_j,u)$, the graph has two arcs: $(\bar{l}_i,l_j)$ and $(\bar{l}_j,l_i)$. We say that these two arcs are {\em complementary}.
\item if the formula contains the unit clause $(l,u)$, we say that vertex $l$ is a {\em starting} vertex, and vertex $\bar{l}$ is an {\em ending} vertex.
\end{itemize}
It is easy to see that if there is a path $(l_1,l_2,\ldots,l_k)$, where $l_1$ and $l_k$ are starting and ending, respectively, and the path does not cross any pair of complementary arcs, then chain resolution can be applied and the path tells the order in which resolution must be applied.

In our implementation, we select one arbitrary starting vertex and compute shortest paths to all ending vertices using Dijkstra's algorithm. If one of the paths does not cross complementary arcs, we trigger the rule. Else, another starting vertex is selected and the process is repeated. Note that this method does not necessarily detect all the potential applications of chain resolution because it only takes into consideration one path between each pair of starting and ending vertices (the shortest path given by Dijkstra). The fact that this path crosses complementary arcs does not prevent the existence of other paths that do not cross complementary arcs. We believe that a better approach would be to use a flow algorithm, but we have not yet studied this possibility.

\section{Experimental Results}
We divide the experiments in two parts. The purpose of the first part is to assess the importance of each one of the inference rules that our solver incorporates. These experiments include random Max-SAT instances and random Max-clique problems. The purpose of the second part is to evaluate the performance of our solver in comparison to other available solving techniques. These experiments include random weighted and unweighted Max-SAT instances, random and structured Max-one problems, random Max-cut problems, random and structured max-clique problems and combinatorial auctions.

Our solver, written in C, is available as part of the {\sc ToolBar} software\footnote{{\tt http://carlit.toulouse.inra.fr/cgi-bin/awki.cgi/ToolbarInfo}}. Benchmarks are also available in the {\sc ToolBar} repository.
In all the experiments with random instances, samples have 30 instances and plots report mean {\em cpu} time in seconds. Executions were made on a 3.2 Ghz Pentium 4 computer with Linux. Unless otherwise indicated, executions were aborted when they reached a time limit of $1200$ seconds. In all the plots' legend, the order of the items reflects the relative performance order of the different competitors.

\subsection{Adding Inference to Max-DPLL}

We consider the following versions of our solver:
\begin{enumerate}
\item Basic Max-DPLL. Namely, Algorithm \ref{f_dpll} in which lines 6-8 in Function \texttt{Simplify} are commented out. We denote this algorithm Max-DPLL-1.
\item The previous algorithm enhanced with {\em neighborhood resolution} (namely, lines 7-8 in \texttt{Simplify} are commented out). We denote this algorithm Max-DPLL-2.
\item The previous algorithm enhanced with {\em chain resolution} (namely, line 8 in \texttt{Simplify} is commented out). We denote this algorithm Max-DPLL-3.
\item The previous algorithm enhanced with {\em cycle resolution} (namely, all the lines in \texttt{Simplify} are considered). We denote this algorithm Max-DPLL-4.
\end{enumerate}

For the first experiment we consider random Max-SAT instances.
A {\em random} $k$-SAT formula is defined by three parameters $<k,n,m>$. $k$ is the length of the clauses, $n$ is the number of variables and $m$ is number of clauses. Each clause is randomly generated by selecting $k$ distinct variables with a uniform probability distribution. The sign of each variable in each clause is randomly decided. In the following experiments we generate instances in which the number of clauses is always sufficiently high as to make the formula unsatisfiable and we solved the corresponding Max-SAT problem. We used the  {\em Cnfgen}\footnote{\texttt{ftp://dimacs.rutgers.edu/pub/challenge/satisfiability/\\contributed/UCSC/instances}} generator. Note that it allows repeated clauses, so $v$ repetitions of a  clause $C$ are grouped into one weighted clause $(C,v)$. 

\begin{figure}[t]
        \centering
	\includegraphics[width=6.5cm]{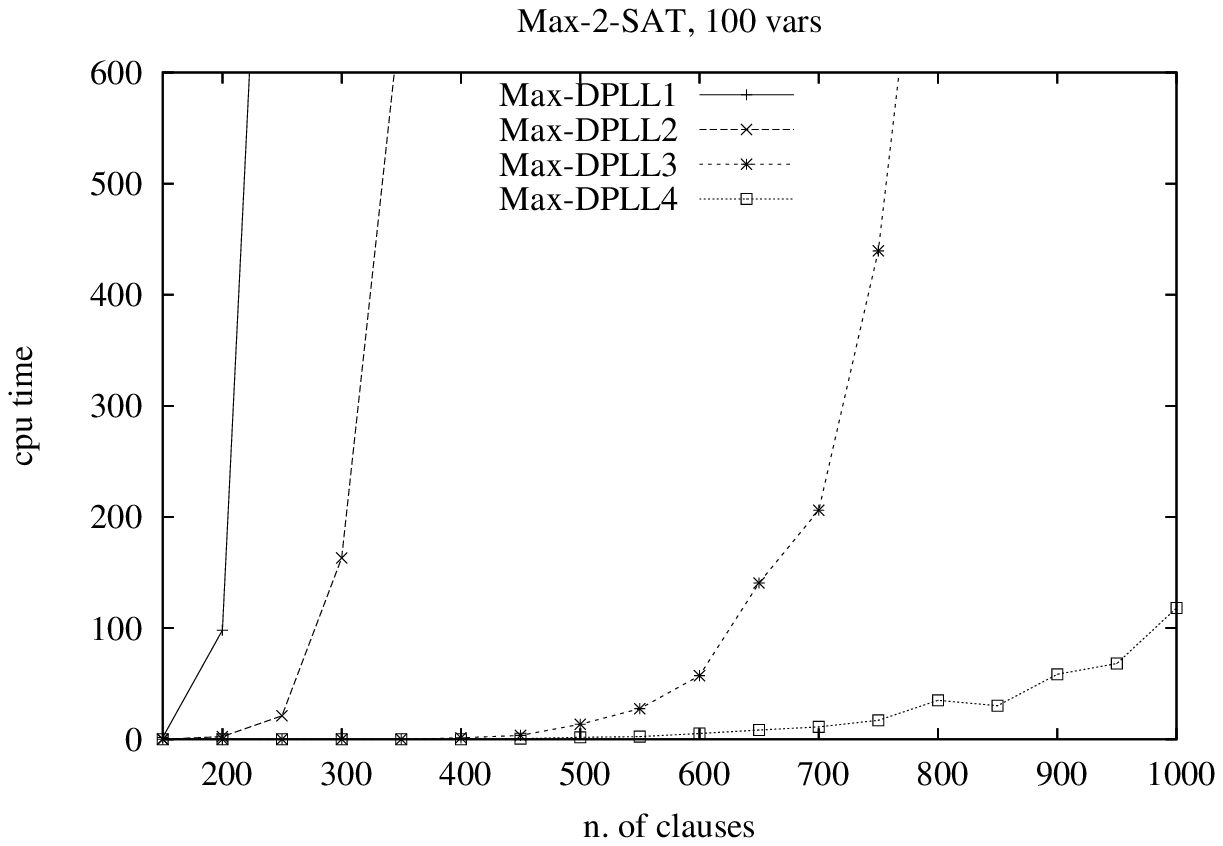}
	\includegraphics[width=6.5cm]{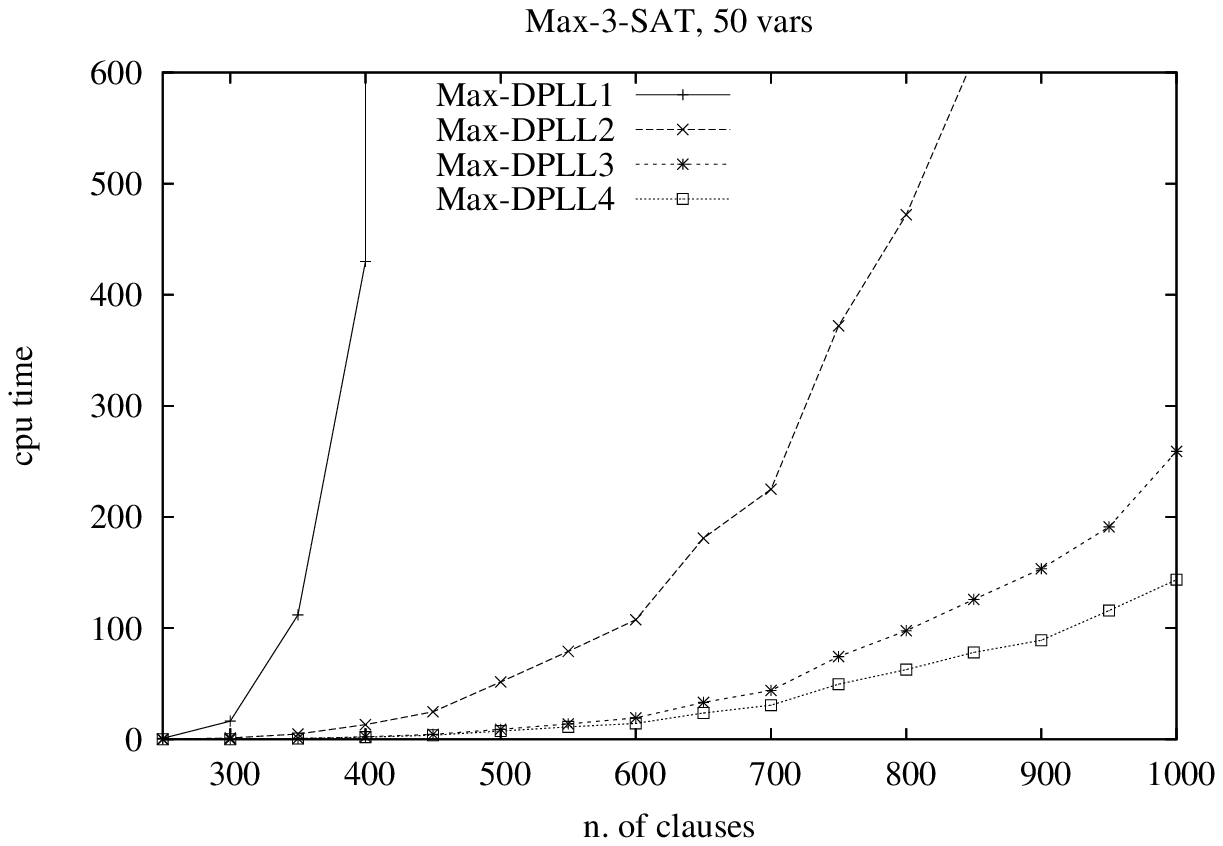}
	\includegraphics[width=6.5cm]{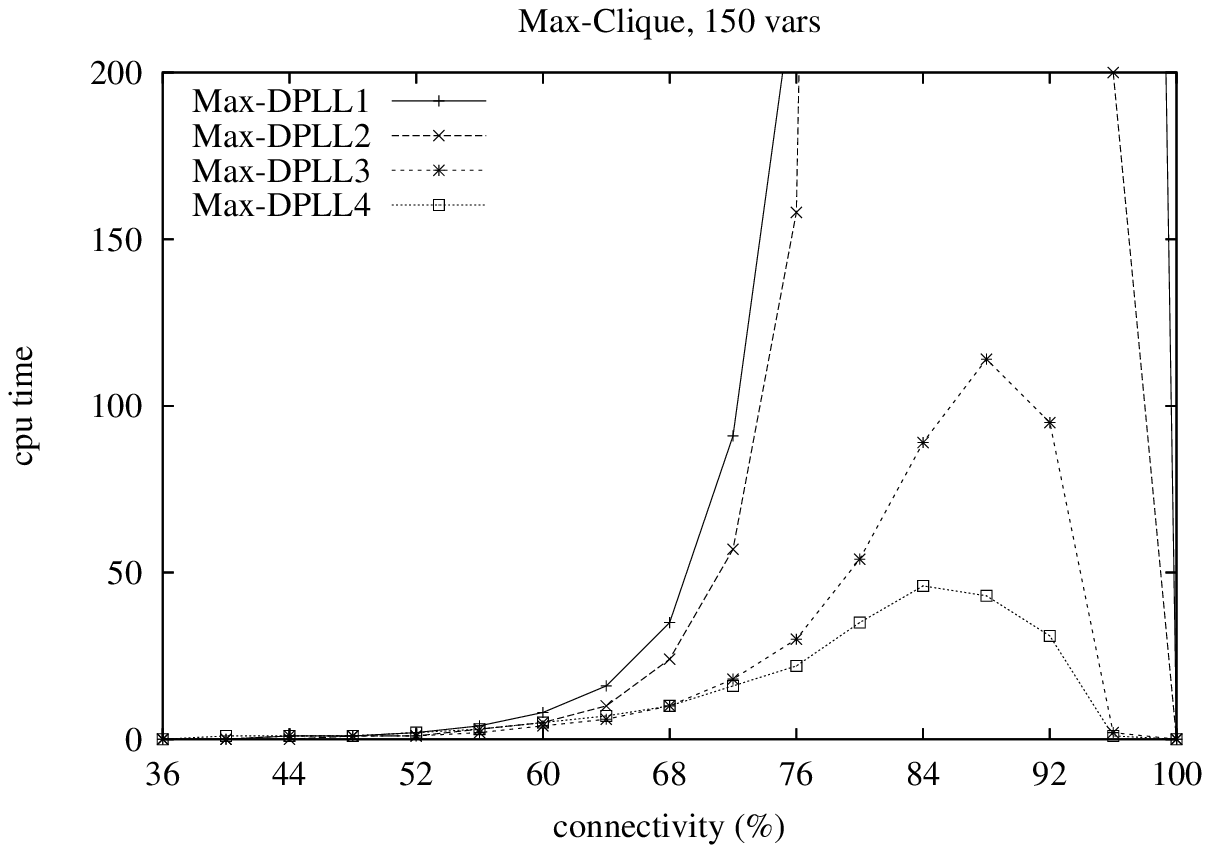}
	\caption{Experimental results of different algorithms on random Max-SAT and Max-clique instances.}
\label{p_exp_alone}
\end{figure}

Figure \ref{p_exp_alone} (top-left) reports results on random Max-2-SAT instances with $100$ variables with varying number of clauses. It can be seen that Max-DPLL-1 performs very poorly and can only solve instances with up to 200 clauses. The addition of neighborhood resolution (namely, Max-DPLL-2) improves its performance by 2 orders of magnitude and allows to solve instances with up to 300 clauses. The further addition of chain resolution provides a spectacular improvement which allows to solve instances with up to 750 clauses. Finally, the addition of cycle resolution allows to solve in $100$ seconds instances of up to $1000$ clauses. 

The Max-Clique problem is the problem of finding the maximum size clique embedded in a given graph. It is known that solving the Max-clique problem of graph $G=(V,E)$ is equivalent to solving the Min-covering problem of graph $G'=(V,E')$ where $E'$ is the complementary of $E$ (namely, $(u,v)\in E'$ iff  $(u,v)\notin E$). Therefore, we solved Max-clique instances by encoding into Max-SAT the corresponding min-vertex problem as described in Example 5.

A {\em random graph} is defined by two parameters $<n,e>$ where $n$ is the number of nodes and $e$ is the
number of edges. Edges are randomly decided using a uniform probability distribution. Figure \ref{p_exp_alone} (bottom) reports the results of solving the max-clique problem of random graphs with $150$ nodes and varying number of edges. It can be observed that the instances with connectivity lower than $50$ percent are trivially solved by our $4$ algorithms. Note that instances with small connectivity have an associated Max-SAT encoding containing a large number of hard clauses. Hence, the {\em unit clause reduction rule} is applied very frequently on those instances. This is the reason why they are so easily solved. However, as the connectivity is increased, the differences between all the versions are greater. Little improvement is noticed for Max-DPLL-2 over Max-DPLL-1. For connectivities between $76 \%$ and $99 \%$ the greatest differences are found. While Max-DPLL-1 and Max-DPLL-2 are unable to solve those instances, both Max-DPLL-3 and Max-DPLL-4 perform well. With a connectivity near to $90 \%$, it can be observed that using the cycle resolution reports a noticeable improvement.

From these experiments we conclude that the synergy of the three inference rules of Max-DPLL-4 produces an efficient algorithm.

\subsection{Max-DPLL versus alternative solvers}

In the following experiments, we evaluate the performance of Max-DPLL-4 (we will refer to it simply as {\sc Max-DPLL}). For that purpose, we compare Max-DPLL with the following state-of-the-art Max-SAT solvers: {\sc Maxsolver} \cite{Zhang05}, {\sc UP} \cite{Planes05} and {\sc LB4a} \cite{Shen04}. They suffer from the following limitations: 
\begin{itemize}
\item  The available version of {\sc Maxsolver} is restricted to instances with less than 200 variables and 1000 clauses.
\item For implementation reasons, {\sc UP} cannot deal with instances having clauses with high weights. Similarly, it cannot also deal with instances that combine mandatory and weighted clauses.
\item {\sc LB4a}  can only solve unweighted Max-2-SAT problems (i.e, it is restricted to binary clauses with unit weights and without repeated clauses).
\end{itemize}
Consequently, in the experiments we will only execute a solver if it is possible, according to its limitations.

It is known that Max-SAT problems can also be solved with {\em pseudo-boolean} and SAT solvers.
For the sake of a more comprehensive comparison, we also consider {\sc Pueblo}  \cite{Hossein06} and {\sc Minisat} \cite{Een03}, which are among the best pseudo-boolean and SAT solvers, respectively. In appendix B, we describe how we translated the Max-SAT instances into these two frameworks.
Note that pseudo-boolean formulas are equivalent to 0-1 {\em integer linear programs} (ILP). Thus, they can also be solved with a state-of-the-art ILP solver such as CPLEX. We have not considered this alternative because \cite{Givry03} showed that it is generally ineffective for Max-SAT instances. Max-SAT problems can also be solved with WCSP solvers \cite{Givry03}. We have not consider this type of solver in our study, because the reference WCSP solver is MEDAC \cite{deGivry05}, which uses techniques similar to those of Max-DPLL and can be roughly described as a non-boolean restricted version of Max-DPLL-3.

\subsubsection{Random Max-k-SAT}

For the following experiment, we generated random 2-SAT instances of 60 variables and 3-SAT instances of 40 variables with varying number of clauses using the {\em Cnfgen} generator. We also generated random 2-SAT instances of 140 variables using the 2-SAT generator of \cite{Shen04} that does not allow repeated clauses. 

\begin{figure}[t]
        \centering
	\includegraphics[width=6.5cm]{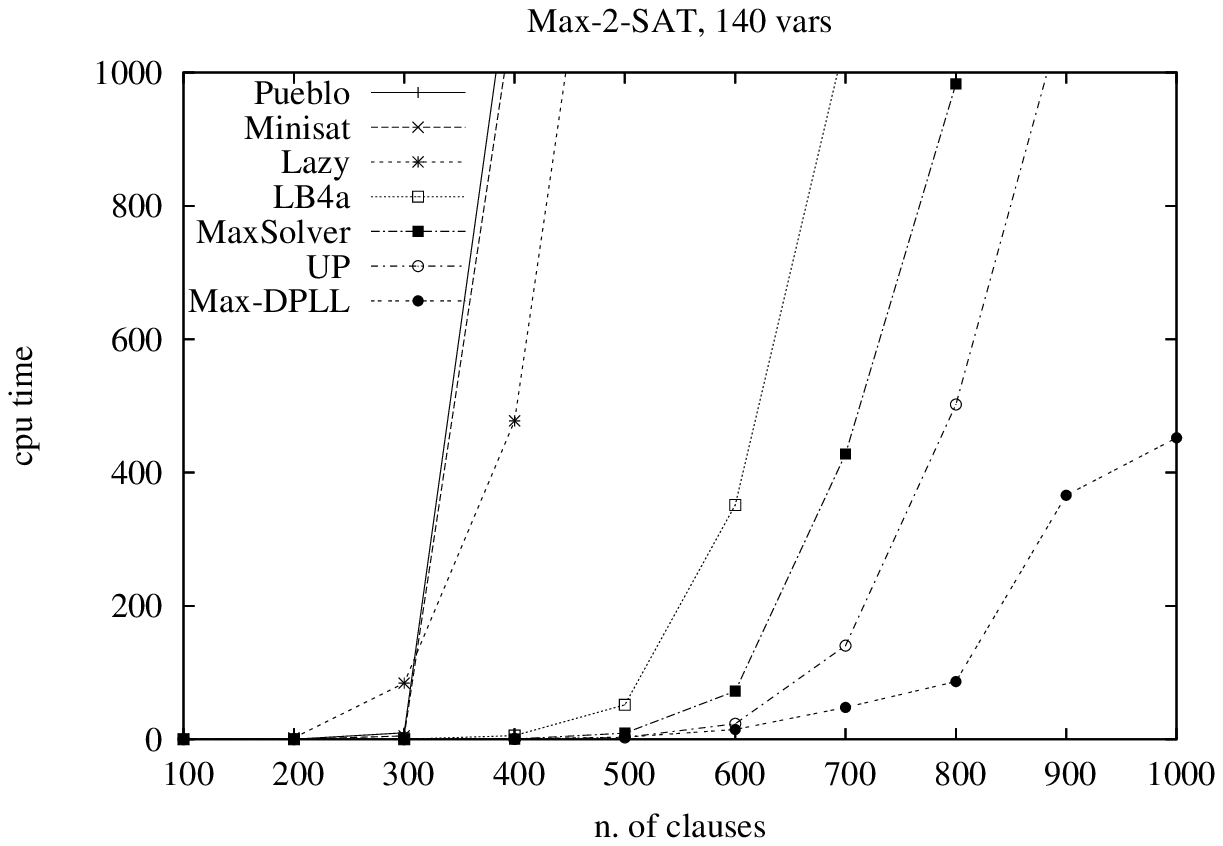}
	\includegraphics[width=6.5cm]{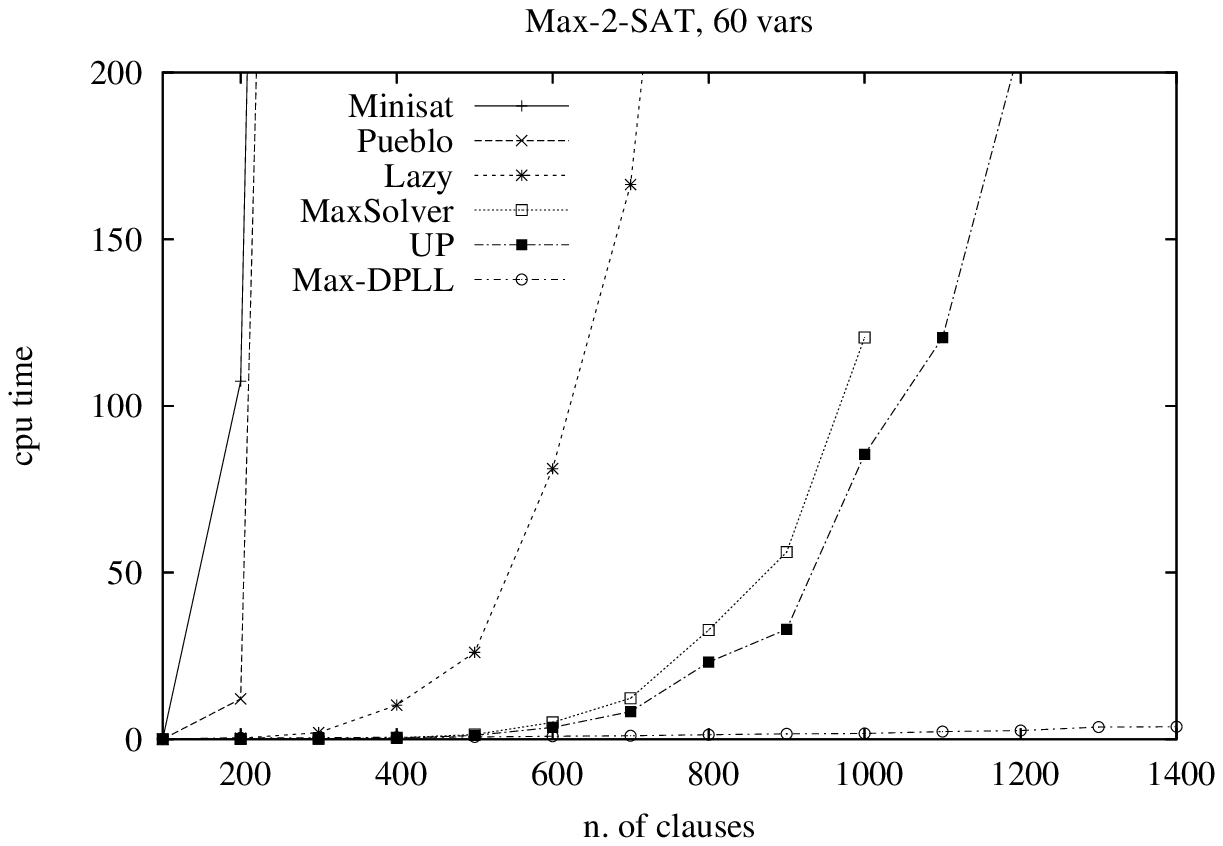}
	\includegraphics[width=6.5cm]{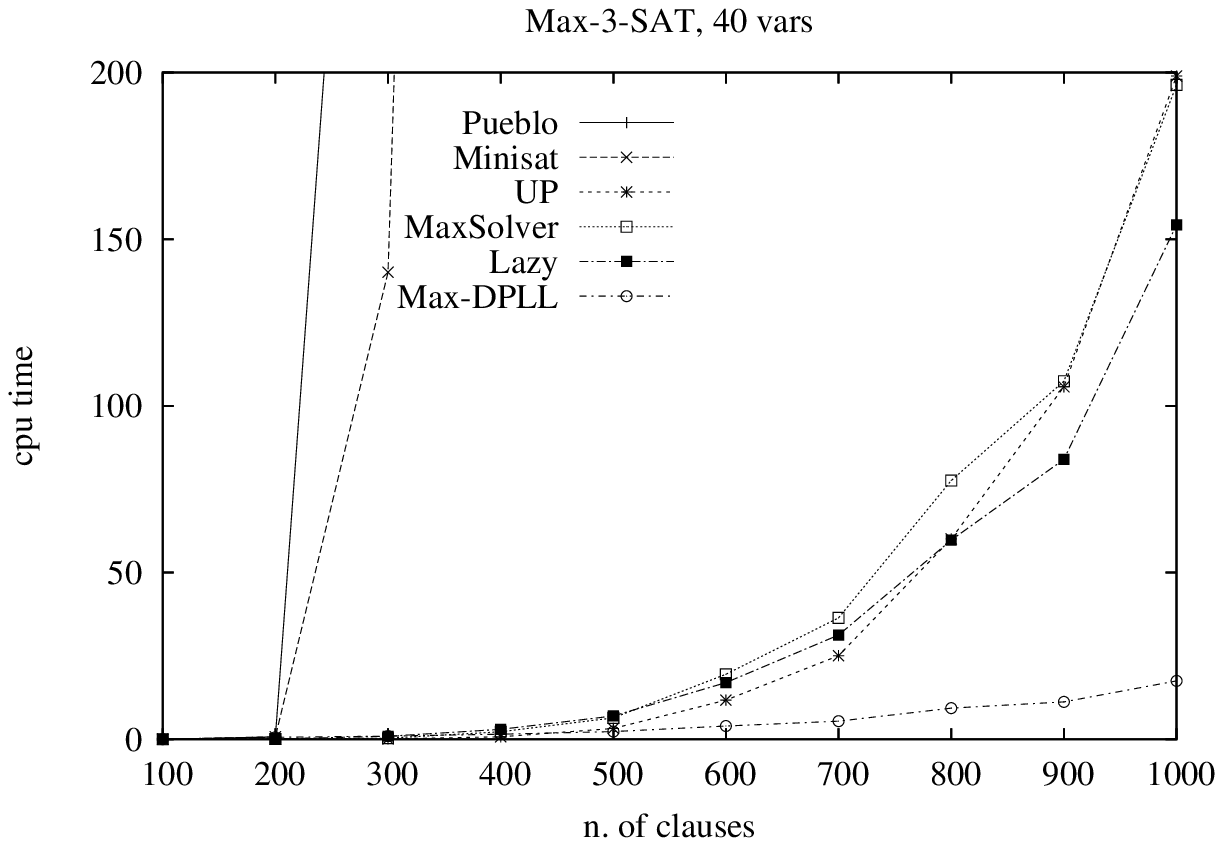}
	\caption{Random Max-2-SAT and Max-3-SAT. Max-2-SAT instances on the plot on the left do not contain repeated clauses.}
\label{p_maxsat_all}
\end{figure}

Figure \ref{p_maxsat_all} (top-left) presents the results on Max-2-SAT without repeated clauses.  It can be observed that {\sc Max-DPLL} is the only algorithm that can solve problems of up to 1000 clauses. The solver with the second best performance, {\sc UP}, is $6$ times slower. A surprising observation is that the {\sc LB4a} solver, which was specifically designed for Max-2-SAT without repetitions, performs worse than the other Max-SAT solvers in random unweighted Max-2-SAT. Figure \ref{p_maxsat_all} (top-right) presents the results on Max-2-SAT with repeated clauses. {\sc Max-DPLL} is again the best algorithm. The second best solver, {\sc UP}, is nearly 100 times slower in the hardest instances.
Figure \ref{p_maxsat_all} (bottom) presents the results on Max-3-SAT. {\sc Max-DPLL} provides again the best performance. The second best option {\sc Lazy} is about $10$ times slower.
A worth noting observation is that the alternative encodings (namely, pseudo-boolean and SAT) do not seem to be effective in these instances.

\subsubsection{Max-one}

Given a satisfiable CNF formula, {\em max-one} is the problem of finding a model with a maximum number of variables set to true. This problem can be encoded as Max-SAT by considering the clauses in the original formula as mandatory and adding a weighted unary clause $(x_i,1)$ for each variable in the formula. Note that solving this problem is much harder than solving the usual SAT problem, because the search cannot stop as soon as a model is found. The optimal model must be found and its optimality must be proved.

\begin{figure}[t]
        \centering
	\includegraphics[width=10cm]{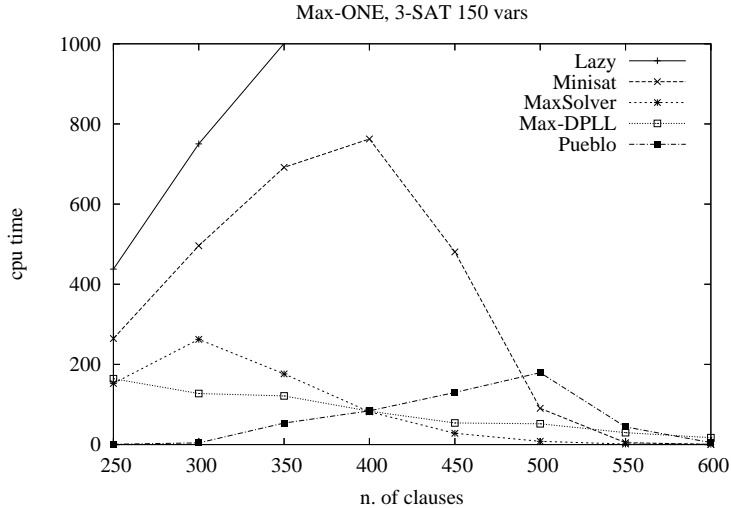}
	\caption{Random Max-one instances.}
\label{p_maxone_1}
\end{figure}

Figure \ref{p_maxone_1} shows results with random 3-SAT instances of $150$ variables. Note that {\sc UP} can not be executed in this benchmark because it cannot deal with mandatory and weighted clauses simultaneously. The first thing to be observed is that {\sc Lazy} and {\sc Minisat} do not perform well. Regarding the other solvers, {\sc Pueblo} is the best when the number of clauses is very small, but its relative efficiency decreases as the number of clauses grows. {\sc MaxSolver} has the opposite behavior, and {\sc Max-DPLL} always lay in the middle. The performance of all these solvers converge as the number of clauses approaches the phase transition peak. The reason is that, as the number of models decreases, the optimization part of the Max-one problem loses relevance (the number of models to chose from decreases).

Table \ref{p_maxone_2} reports results on the Max-one problem on selected satisfiable SAT instances from the DIMACS challenge. The first column indicates the name of the problem classes. The second column indicates the number of instances of each class. The rest of columns indicate the performance of each solver by indicating the number of instances that could be solved within the time limit. If all the instances could be solved, the number in parenthesis is the mean time in seconds. The ``-'' symbol in the {\sc MaxSolver} column indicates that the instances could not be executed due to the limitation that this solver has on the maximum number of variables and clauses. As can be observed, {\sc MaxSolver} and {\sc Lazy} do not succeed in this benchmark, which means that {\sc Max-DPLL} is the only Max-SAT solver that can deal with it. Its performance is comparable to the good performance of {\sc Minisat} and {\sc Pueblo}. However, in the {\em par16*c*} instances  {\sc Max-DPLL} performs badly, while in the {\em par8*} instances it performs better than the others.

\begin{figure}[t]
\begin{center}
\begin{tabular}{c|c|c|c|c|c|c|c|c}
Problem & n. inst. &  MaxDPLL & MaxSolver & Lazy & Minisat & Pueblo  \\
\hline
aim50* & $16$    &  $16 (0.59)$ &  $16 (0.12)$ &  $16 (28.25)$ & $16 (0.01)$ & $16 (0.00)$\\
aim100* & $16$   &  $16 (2.67)$ &  $16 (4.92)$ &  $0$ & $16 (0.02)$ & $16 (0.00)$\\
aim200* & $16$    &  $9$ &  $4$ &  $0$ & $16 (0.03)$ & $16 (0.00)$\\
jnh* & $16$    &  $16 (1.49)$ & $-$ &  $6$ & $16 (0.08)$ & $16 (0.10)$\\
ii8* & $14$    &  $5$ & $-$ &  $1$ & $10$ & $3$\\
ii32* & $17$    &  $11$ &  $-$ &  $0$ & $16$ & $15$\\
par8* & $10$    &  $10 (0.92)$ & $-$ &  $5$ & $10 (16.39)$ & $10 (26.52)$\\
par16*c* & $5$    &  $5 (784.14)$ & $-$ &  $0$ & $5 (0.93)$ & $5 (0.93)$\\
\end{tabular}
\end{center}
\caption{Results for the Max-one problem on selected DIMACS SAT instances.}
\label{p_maxone_2}
\end{figure}

\subsubsection{Max-cut}

Given a graph $G=(V,E)$, a {\em cut} is defined by a subset of vertices $U\subseteq V$. The size of a cut is the number of edges $(v_i,v_j)$ such that $v_i\in U$ and $v_j\in V-U$. The {\em Max-cut} problem consists on finding a cut of maximum size. It is encoded as Max-SAT associating one variable $x_i$ to each graph vertex. Value $\mathbf{t}$ (respectively, $\mathbf{f}$) indicates that vertex $v_i$ belongs to $U$ (respectively, to $V-U$). For each edge $(v_i,v_j)$, there are two clauses $x_i \vee x_j, \bar{x}_i \vee \bar{x}_j$. Given a complete assignment, the number of violated clauses is $|E|-S$ where $S$ is the size of the cut associated to the assignment. Note that this encoding produces an unweighted Max-$2$-SAT formula, so the {\sc LB4a} solver can be used. Random Max-Cut instances are extracted from random graphs. We considered graphs of 60 nodes with varying number of edges.

\begin{figure}[t]
        \centering
	\includegraphics[width=10cm]{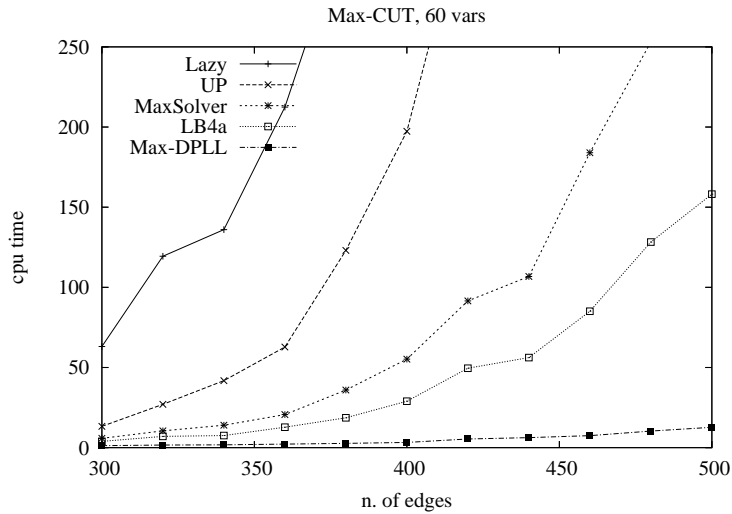}
	\caption{Random Max-cut instances.}
\label{p_maxcut}
\end{figure}

Figure \ref{p_maxcut} reports the results on this benchmark. It can be observed that for all solvers other than {\sc Max-DPLL}, problems become harder as the number of edges increases. However, {\sc Max-DPLL} solves instances of up to $500$ edges almost instantly. The second best solver is {\sc LB4a}, but {\sc Max-DPLL} is  up to $15$ times faster.
{\sc Pueblo} and {\sc Minisat} perform so poorly even in the easiest instances that they are not included in the comparison.

\subsubsection{Max-clique}

\begin{figure}[t]
        \centering
	\includegraphics[width=10cm]{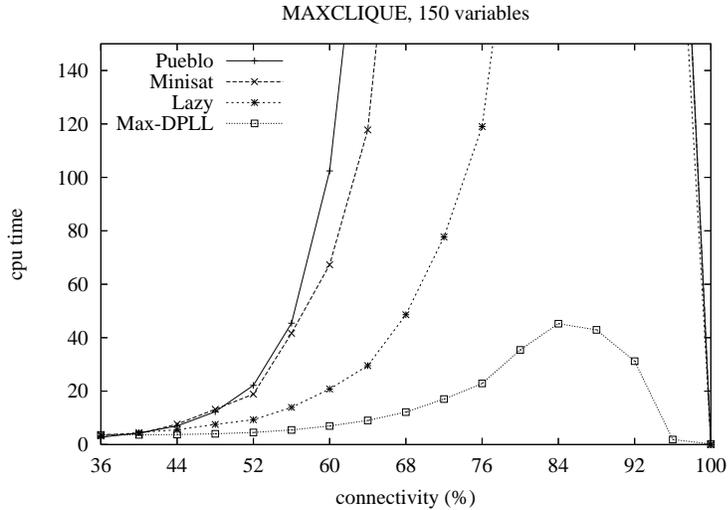}
	\caption{Random Max-clique instances.}
\label{p_maxclique_1}
\end{figure}

The Max-clique problem is the problem of finding the maximum size subgraph embedded in a given graph and its Max-SAT encoding was described in the previous subsection. Solvers {\sc UP}, {\sc MaxSolver} and {\sc Lb4a} could not be executed in this domain due to their limitations.
Our first Max-clique experiment used random graphs with $150$ nodes and varying number of edges.
Figure \ref{p_maxclique_1} reports the results. Again, {\sc Max-DPLL} is clearly better than any other competitor. All other competitors are more than 2 orders of magnitude slower than {\sc Max-DPLL}.

%{\sc UP} could not be executed in this benchmark because it cannot deal with mandatory and weighted clauses.

We also considered the $66$ Max-Clique instances from the DIMACS challenge\footnote{\texttt{ftp://dimacs.rutgers.edu/pub/challenge/graph/benchmarks/clique}}. {\sc MaxSolver} could not be executed in this benchmark because the number of variables and clauses of the instances exceeds its capacity. Thus, the only two Max-SAT solvers that could be executed are {\sc Max-DPLL} and {\sc Lazy}. Within the time limit, they solved 32 and 23 instances, respectively. {\sc Minisat} and {\sc Pueblo} could solve 22 and 16 instances, respectively. Therefore, {\sc Max-DPLL} provided the best performance in this benchmark, too.

These instances have been previously used to evaluate several dedicated max clique algorithms.
Performing a proper comparison with {\sc Max-DPLL} is difficult because their code is not available and we would need to re-program their algorithms. However, following the approach of \cite{Reg03}, we overcome this problem by normalizing the reported times. Of course, this is a very simplistic approach which disregards very relevant parameters such as the amount of memory or the processor model. In consequence, the following results can only be taken as orientative. Giving a time limit of $2.5$ hours per instance in our $3.2$ Ghz computer, {\sc Max-DPLL} was able to solve $37$ instances. In an {\em equivalent} (via normalization) time, \cite{Wood97} solves 38, \cite{Oster02} soves 36, \cite{Fahle02} solves 45, and \cite{Reg03} solves 52.

\subsubsection{Combinatorial Auctions}

{\em Combinatorial auction}  allow bidders to bid for indivisible subsets of goods. Consider a set $G$ of goods and $n$ bids. Bid $i$ is defined by the subset of requested goods $G_i \subseteq G$ and the amount of money offered. The bid-taker, who wants to maximize its revenue,  must decide which bids are to be accepted. Note that if two bids request the same good, they cannot be jointly accepted \cite{sandholm99}. In its Max-SAT encoding, there is one variable $x_i$ associated to each bid. There are unit clauses $(x_i,u_i)$ indicating that if bid $i$ is not accepted there a loss of profit $u_i$. Besides, for each pair $i,j$ of conflicting bids, we add a mandatory clause $(\bar{x_i}\vee \bar{x_j},\top)$.

We used the CATS generator \cite{Leyton00} that allows to generate random instances inspired from real-world scenarios. In particular, we generated instances from the {\em Regions}, {\em Paths} and {\em Scheduling} distributions. The number of goods was fixed to $60$ and we increased the number of bids. By increasing the number of bids, instances become more constrained (namely, there are more conflicting pairs of bids) and harder to solve. {\sc UP}, {\sc MaxSolver} and {\sc Lb4a} could not be executed due to their limitations. The {\sc Lazy} solver could not be included in the {\em Regions} comparison due to overflow problems.

\begin{figure}[t]
        \centering
	\includegraphics[width=6.5cm]{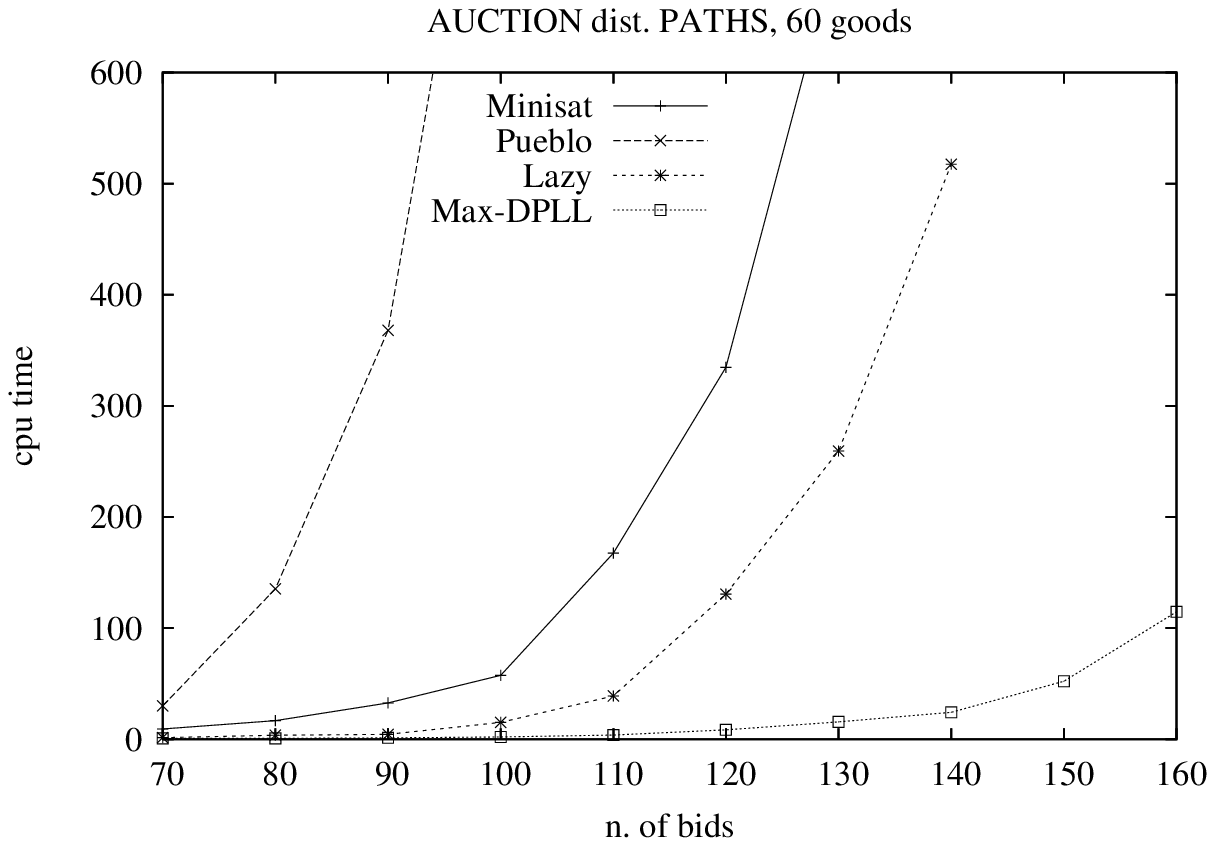}
	\includegraphics[width=6.5cm]{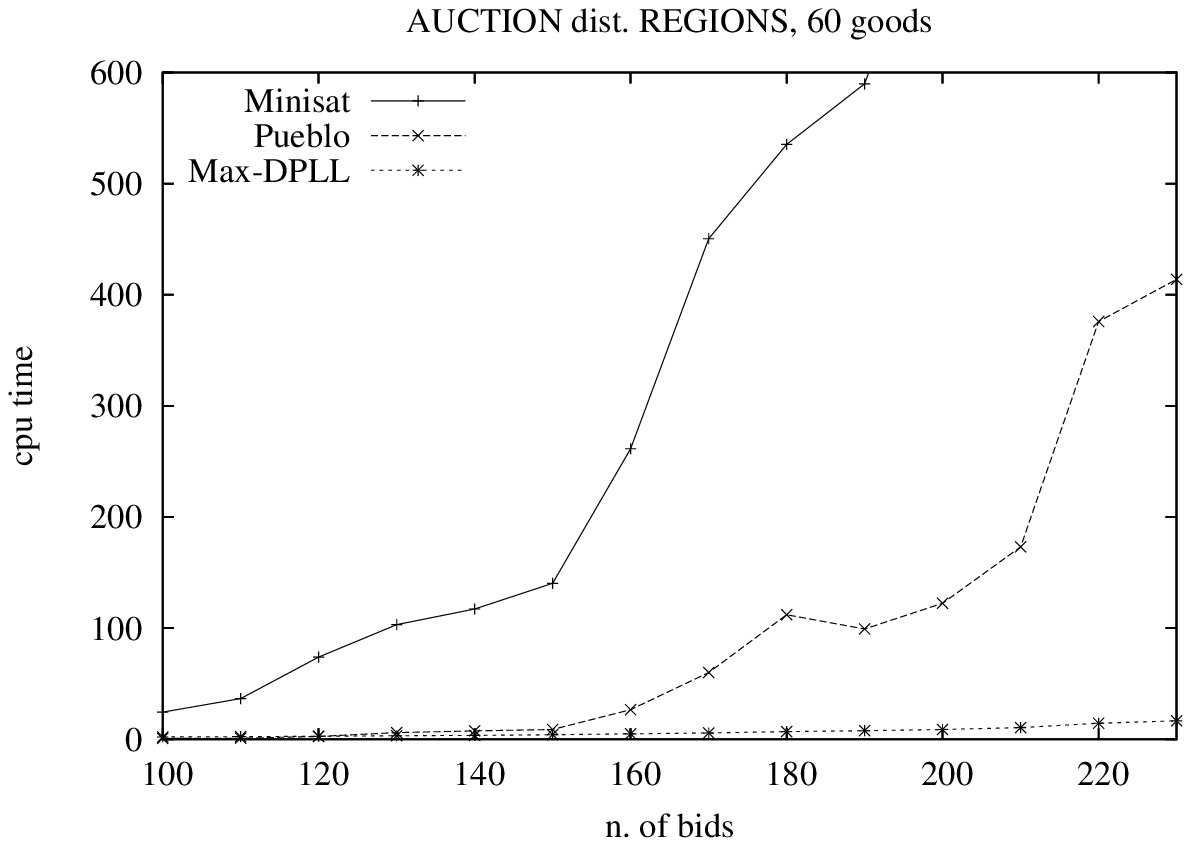}
	\includegraphics[width=6.5cm]{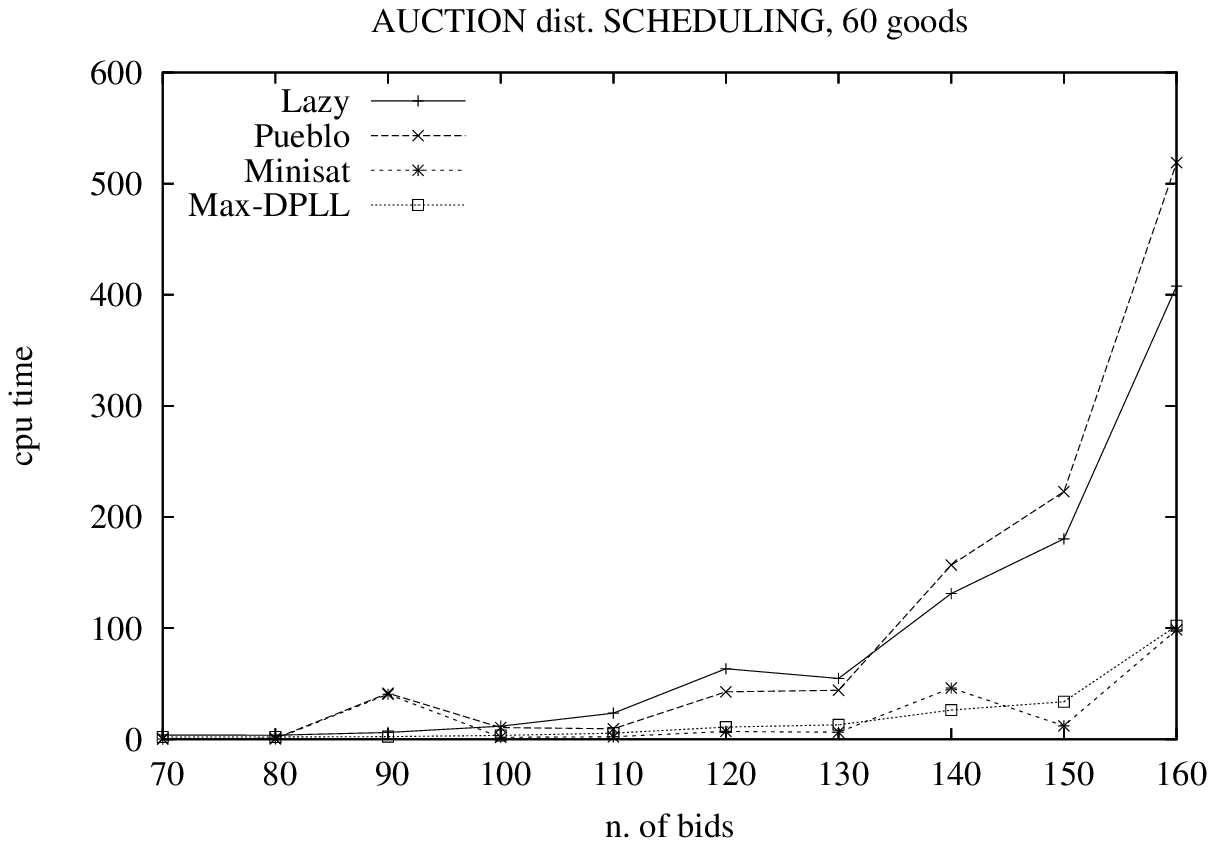}
	\caption{Combinatorial auctions. Top-left: {\em Paths} distribution. Top-right: {\em Regions} distribution. Bottom: {\em Scheduling} distribution.}
\label{p_ca}
\end{figure}

Figure \ref{p_ca} (top-left) presents the results for the Paths distribution. {\sc Max-DPLL} produces the best results being $22$ times faster than the second best option {\sc Lazy}. Figure \ref{p_ca} (top-right) presents the results for the Regions distribution. {\sc Max-DPLL} is again the best algorithm. It is $26$ times faster than the second best solver {\sc Pueblo}. Finally, results for the Scheduling distribution are shown in Figure \ref{p_ca} (bottom). In this benchmark, the performance of {\sc Max-DPLL} and {\sc Minisat} are quite similar, while the other solvers are up to $4$ times slower.

\section{Conclusions and Future work}

This paper introduces a novel Max-SAT framework which highlights the relationship between SAT and
Max-SAT solving techniques. Most remarkably, it extends the concept of {\em resolution}. Our resolution rule, first proposed in \cite{Larrosa05a}, has been proved complete in \cite{Bonet06}. There are many beneficial consequences of this approach:
\begin{itemize}
\item It allows to talk about Max-SAT solving with the usual SAT terminology.
\item It allows to naturally extend basic algorithms such as DPLL and DP.
\item It allows to express several solving techniques that are spread around the Max-SAT literature with a common formalism, see their logical interpretation and see the connection with similar SAT, CSP and WCSP techniques.
\end{itemize}

From a practical point of view, we have proposed a hybrid algorithm that combines search and selected forms of inference. It follows a typical search strategy but, at each visited node, it attempts to simplify the current subproblem using special cases of resolution with which the problem is transform into a simpler, equivalent one. Our experiments on a variety of domains show that our algorithm is usually orders of magnitude faster than its competitors.

Our current solver lacks features that are considered very relevant in the SAT context (for example clause learning, re-starts, etc). Since our framework makes the connection between SAT and Max-SAT very obvious, they should be easily incorporated in the future. Additionally, some of the ideas presented in this paper have been borrowed from the {\em weighted CSP} field \cite{Meseguer06}. Therefore, it seems also possible to incorporate new (weighted) constraint processing techniques. Finally, we want to note the recent work of \cite{Planes05} in which very good lower bounds are obtained by temporarily setting $\top=1$ and simulating unit propagation. Since the hyper-resolution rules presented in Section 5.2 are special cases of their more general algorithm, we want to explore if their approach can be fully described with our resolution rule.

\appendix

\section{Correctness and Complexity of Max-VarElim}

In this appendix we prove Lemmas 19 and 21, which establish the correctness of the \texttt{Max-VarElim} function in Figure \ref{MaxDP} and its time and space complexity. In the proofs we borrow some ideas from \cite{Rish00,Davis60,Bonet06} and adapt them to our framework.

In the following, when we write $C\in \mathcal{F}$ we mean
$(C,u)\in \mathcal{F}$ for some weight $u$ (there is no ambiguity because all clauses in $\mathcal{F}$ are different). 
We use symbol $\mathcal{F}\vdash_{x_i} \mathcal{F}'$ to denote the application of a resolution step to formula $\mathcal{F}$ resulting in formula $\mathcal{F}'$, where the clashing variable was $x_i$. Consider
the elimination of variable $x_i$ with Function \texttt{Max-VarElim}. First of all, the formula is partitioned into two sets of clauses, $\mathcal{B}$ and $\mathcal{F}$. Then, clauses of the form $(x_i\vee A, u)$ are fetched from $\mathcal{B}$, resolved with clashing clauses until quiescence or disappearance and, finally, are discarded. Suppose that discarded clauses are stored in a set $\mathcal{D}$. Formally, we can see
the execution of \texttt{Max-VarElim} as a sequence of resolution steps,
$$\mathcal{B}_o \cup \mathcal{F}_o \cup \mathcal{D}_o \vdash_{x_i} \mathcal{B}_1 \cup \mathcal{F}_1 \cup \mathcal{D}_1 \vdash_{x_i} \ldots \vdash_{x_i} \mathcal{B}_q \cup \mathcal{F}_q \cup \mathcal{D}_q$$
where $\mathcal{D}_o=\varnothing$. For all $0\leq k \leq q$: $\mathcal{B}_k$ is a set of clauses that contain either $x_i$ or $\bar{x}_i$, $\mathcal{F}_k$ is a set of clauses that {\em do not} contain $x_i$ neither $\bar{x}_i$, and $\mathcal{D}_k$ is a set of clauses that contain $x_i$. Besides, $\mathcal{B}_q$ does not have any clause with $x_i$. The output of \texttt{Max-VarElim} is $\mathcal{F}_q$ that, as we will prove, is essentially equivalent to the original formula.
Let $N_i$ denote the set of variables sharing clauses with $x_i$ in the starting $\mathcal{B}$ (namely, $\mathcal{B}_o$),
$$N_i=\{x_j\neq x_i|\ \ \exists_{C\in \mathcal{B}_o} \ \ x_j\in var(C)\}$$
and let $n_i=|N_i|$ be its cardinality. 
In the remaining of this appendix we will show that: the number of new clauses generated during the sequence of resolution steps is bounded by $O(3^{n_i})$ (space complexity), the number of resolution steps is bounded by $O(9^{n_i})$ (time complexity) and, from an optimal model of $\mathcal{F}_q$ we can be trivially generate an optimal model of the original formula $\mathcal{B}_o \cup \mathcal{F}_o$ (correctness).

Observe that all the variables different from $x_i$ appearing in clauses generated by the resolution process must belong to $N_i$ because resolution does not add new variables. Therefore, all the clauses in $\mathcal{B}_k$ have the form $l\vee A$ where $var(l)=x_i$ and $var(A)\subseteq N_i$. Variable $x_i$ must appear in the clause either as a positive or negative literal (namely, there are 2 options) and every $x_j\in N_i$ may or may not appear in $A$ and, if it appears, it can be in positive or negative form (namely, there are 3 options). Consequently, the size of $\mathcal{B}_k$ is bounded by $2\times 3^{n_i}$. For similar reasons, every clause $C$ {\em added} to $\mathcal{F}$ during the resolution process satisfies that $var(C)\subseteq N_i$. Every $x_j\in N_i$ may or may not appear in $C$ and, if it appears, it may be positive or negative (namely, there are 3 options). Consequently, the number of non-original clauses in $\mathcal{F}_k$ is bounded by $3^{n_i}$. Therefore, the number of clauses added to $\mathcal{B}$ and $\mathcal{F}$ during the execution of \texttt{Max-VarElim} is bounded by 
$2\times 3^{n_i} + 3^{n_i}$. As a result, its space complexity is $O(3^{n_i})$.

Next, we analyze the time complexity. Recall that two clauses  $(x_i\vee A,u), (\bar{x}_i\vee B,w)\in \mathcal{F}$
{\em clash} if $A\vee B$: is not a tautology (i.e., $\forall_{l\in A}\ \bar{l}\notin B$) and, $A\vee B\in \mathcal{F}$ is not absorbed (i.e, $\forall_{(C,\top)\in \mathcal{F}}\ C \not\subseteq A\vee B$). 
We say that a clause $(x_i\vee A,u)$ is {\em saturated} if there is no clause in $\mathcal{F}$ clashing with it. 
The following lemma shows that resolving on a clause, either removes the clause from the formula or reduces the number of clauses clashing with it,

\begin{lem}
Consider a resolution step $\mathcal{P}\vdash_{x_i} \mathcal{P}'$ where $(x_i\vee A,u)$ and $(\bar{x}_i\vee B,w)$ are the clashing clauses. Then, either  $x_i\vee A \notin \mathcal{P}'$ or the number of clauses clashing with $x_i\vee A$ decreases.
\end{lem}
\begin{pf}
We reason by cases:
\begin{enumerate}
\item If $u<w$ or $u=w<\top$ then the posterior $x_i\vee A$ has weight $0$ (namely, disappears from the formula).
\item If $u=w=\top$ then the effect of resolution is to add the resolvent to the formula ($\mathcal{P}'= \mathcal{P} \cup (A\vee B,\top)$). Then, $\bar{x}_i\vee B$ does not clash with $x_i\vee A$ anymore. 
\item If $u>w$ then $\bar{x}_i\vee B$ is replaced by $\bar{x}_i\vee C\vee \bar{A}$ in the formula. The new clause does not clash with $x_i\vee A$, because $A\vee B\vee \bar{A}$ is a tautology.
\end{enumerate}
\end{pf}

Consider the inner loop of \texttt{Max-VarElim}. It selects a clause $x_i\vee A$ and resolves it until either it disappears or it saturates. If $x_i\vee A$ saturates, it is removed from $\mathcal{B}$ and added to $\mathcal{D}$. We call this sequence of resolution steps the {\em processing} of $x_i\vee A$ and use symbol $\vdash_{x_i\vee A}^*$ to represent it. A consequence of the previous lemma is that the number of resolution steps required to process $x_i\vee A$ is bounded by the number of clauses clashing with it. Note that the number of clauses clashing with $(x_i\vee A,u)$ is bounded by $3^{n_i}$, because clashing clauses must belong to $\mathcal{B}$ and variable $x_i$ must occur negated. Therefore, for each iteration of the outer loop, the inner loop of \texttt{Max-VarElim} iterates at most $3^{n_i}$ times.

Consider now the outer loop of \texttt{Max-VarElim}. It selects a sequence of clauses $x_i\vee A_1, x_i\vee A_2, \ldots, x_i\vee A_s$ and processes them one after another. We can see this process as,
$$\mathcal{B}_o \cup \mathcal{F}_o \cup \mathcal{D}_o \vdash_{x_i\vee A_1}^* \mathcal{B}_{k_1} \cup \mathcal{F}_{k_1} \cup \mathcal{D}_{k_1} \vdash_ {x_i\vee A_2}^* \ldots \vdash_{x_i\vee A_s}^* \mathcal{B}_{k_s} \cup \mathcal{F}_{k_s} \cup \mathcal{D}_{k_s}$$
Recall that the algorithm always selects for processing a clause $x_i\vee A_j$
of minimal size (line 4). Observe that the size of  the compensation clause $x_i\vee A \vee \bar{B}$ added to $\mathcal{B}$ (line 9) is larger than the clause that is being processed. As a consequence, once a clause is processed, it does not appear again in $\mathcal{B}$, which means that
$\forall_{1\leq j<j' \leq s}\ A_j\neq A_{j'}$. A direct consequence is that, since there are at most $3^{n_i}$ distinct $A_j$, the outer loop iterates at most $3^{n_i}$. Therefore, the maximum number of iterations of the inner loop is $3^{n_i}\times 3^{n_i}=9^{n_i}$, which means that the time complexity of the function is $O(9^{n_i})$.

Finally, we prove the correctness of \texttt{Max-VarElim}.
\begin{lem}
A saturated clause, remains saturated during any sequence of resolution steps $\vdash_{x_i}$.
\end{lem}
\begin{pf}
Consider a resolution step $\mathcal{F}\vdash_{x_i}\mathcal{F}'$. Let $x_i\vee A$ and $\bar{x}_i\vee B$ be the clashing clauses, and let $x_i\vee C$ be a saturated clause of $\mathcal{F}$. We only need to prove that $x_i\vee C$ remains saturated in $\mathcal{F}'$. Since, $x_i\vee C$ is saturated in $\mathcal{F}$, either $C\vee B$ is a tautology or it is absorbed in $\mathcal{F}$.
The only new clause in $\mathcal{F}'$ that could clash with $x_i\vee C$ is $\bar{x}_i\vee B \vee \bar{A}$. However, if $C\vee B$ was a tautology, so it is $C\vee B \vee \bar{A}$. If $C\vee B$ was absorbed in $\mathcal{F}$, so it will $C\vee B \vee \bar{A}$ in $\mathcal{F}'$.
\end{pf}

A consequence of the previous lemma is that at the end of the sequence of resolution steps performed by \texttt{Max-VarElim} we have a formula
$\mathcal{B}_{k_s} \cup \mathcal{F}_{k_s} \cup \mathcal{D}_{k_s}$ such that all its clauses are saturated.

To prove the correctness of \texttt{Max-VarElim}
we only need to prove that any assignment $I$ of $\mathcal{F}_{k_s}$ can be extended to variable $x_i$ in a cost free-manner, taking into account the clauses $\bar{x}_i\vee B \in \mathcal{B}_{k_s}$ and the clauses $x_i\vee A \in \mathcal{D}_{k_s}$, because it means that finding the optimal assignment of $\mathcal{F}_{k_s}$ is equivalent to finding the optimal assignment of $\mathcal{B}_{k_s} \cup \mathcal{F}_{k_s} \cup \mathcal{D}_{k_s}$ which, in turn is equivalent to finding the optimal assignment of the original formula.

%What follows is a combination of the proofs of \cite{Davis60} and \cite{Bonet06}.
If $\mathcal{B}_{k_s}=\varnothing$ (resp. $\mathcal{D}_{k_s}=\varnothing$), variable $x_i$ must be set to {\em true} (resp. {\em false}). Else, consider that there is a clause $x_i\vee A\in \mathcal{D}_{k_s}$ such that $I$ does not satisfy $A$ (similarly for $\bar{x}_i \vee B \in \mathcal{B}_{k_s}$). Variable $x_i$ must be set to {\em true}. We show that $I$ satisfies every $\bar{x}_i\vee B\in \mathcal{B}_{k_s}$: Clause $x_i\vee A$ is saturated, then either $A\vee B$ is a tautology or there is a clause $C\in \mathcal{F}_{k_s}$ with $C\subseteq A\cup B$. In the first case, since $I$ does not satisfy $A$, and since $A\vee B$ is a tautology, this means that $I$ satisfies $B$.
In the second case, since $I$ satisfies $C$ and does not satisfy $A$, it must satisfy $B$.

\section{Solving Max-SAT with Pseudo-boolean and SAT solvers}

In {\em Linear pseudo-Boolean} (LPB) problems  over boolean variables $\{x_1,\ldots,x_n\}$, values {\em true} and {\em false} are replaced by numbers $1$ and $0$, respectively. 
Literal $l_i$ represents either $x_i$ or its {\em negation} $1-x_i$.
A LPB problem is defined by a LPB objective function (to be minimized),
$$\sum_{i=1}^n a_{i} l_i \ \ \textrm{where}\ a_{i} \in \mathbb{Z}$$
and a set of LPB constraints,
$$\sum_{i=1}^n a_{ij} l_i \geq b_j,\ \ \textrm{where}\ a_{ij},b_j, \in \mathbb{Z},\ \ x_i\in \{0,1\}$$

A Max-SAT formula can be encoded as a LPB problem \cite{Givry03} by partitioning the set of clauses into three sets: $\mathcal{H}$ contains the mandatory clauses $(C,\top)$, $\mathcal{W}$ contains the non-unary non-mandatory clauses $(C,u<\top)$ and $\mathcal{U}$ contains the unary non-mandatory clauses $(l,u)$. 
For each hard clause $(C_j,\top)\in \mathcal{H}$ there is a LPB constraint $C'_j \geq 1$, where $C'_j$ is obtained from $C_j$ by replacing $\vee$ by $+$ and negated variables $\bar{x}$ by $1-x$. 
For each non-unary weighted clause $(C_j,u_j)\in \mathcal{W}$ there is a LPB constraint $C'_j+r_j \geq 1$, where $C'_j$ is computed as before, and $r_j$ is a new variable that, when set to $1$, trivially satisfies the constraint.
Finally, the objective function is,
$$\sum_{(C_j,r_j)\in \mathcal{W}} u_j r_j + \sum_{(l_j,u_j) \in \mathcal{U}} u_j l_j \geq \top$$

A LPB problem can be solved with a native LPB solver such as {\sc Pueblo} or with a SAT solver. In the latter case, each LPB constraint must be converted into a logic circuit. There are different possible conversions such as BDDs, adders or sorters. In our experiments we used MiniSAT+ \cite{Niklas06}, a translating tool that converts each PB constraint into the presumably more convenient circuit and solves the corresponding SAT formula with {\sc MiniSAT}. {\sc MiniSAT+} converts the objective function of the LPB problem into another LPB constraint by setting an upper bound. The LPB problem is solved by decreasing the value of the upper bound until an infeasible SAT formula is found.

\shrink{
A {\em Weighted constraint satisfaction problem} (WCSP) has a set of variables $Y=\{y_1,\ldots,y_n\}$, each one with a finite set $D_i$ of feasible values, and a set of cost functions $\{f_1,\ldots,f_e\}$. Each cost function $f_j$, defined over a subset of the variables $Y^j\subseteq Y$, associates a natural number from the interval $[0..\top]$ to each possible assignment of $Y^j$. The objective function is,
$$\min_Y \sum_{i=1}^e f_i(Y^i)$$
A Max-SAT formula can be encoded as a WCSP problem as follows \cite{Givry03}: The boolean variables are the WCSP variables. Each set of clauses $\{(C_{i_1},u_{i_1}), (C_{i_2},u_{i_2}),\ldots, (C_{i_r},u_{i_r})\}$ with the same scope 
$var(C_{i_1})=var(C_{i_2})=\ldots=var(C_{i_r}) =Y^j$ define a cost function,

\[f_i(Y^i) = \left\{ \begin{array}
            {r@{\quad:\quad}l}
            u_j & Y^i\ \mathrm{falsifies}\ C_{i_j} \\
            0 & \mathrm{in other case}
            \end{array} \right.
\]

}

\bibliography{../../general}
\bibliographystyle{elsart-num}
\end{document}